\documentclass{article}

\PassOptionsToPackage{numbers,compress}{natbib}
\usepackage[main,final]{neurips_2026}
\usepackage{caption}
\usepackage{graphicx}
\usepackage{subcaption}
\usepackage[utf8]{inputenc} 
\usepackage[T1]{fontenc}    
\usepackage{hyperref}\hypersetup{hidelinks}       
\usepackage{url}            
\usepackage{booktabs}       
\usepackage{amsfonts}       
\usepackage{nicefrac}       
\usepackage{microtype}      
\usepackage{xcolor}         
\newcommand{\oursolution}{CasePlay}
\usepackage{multirow}
\usepackage{colortbl}
\usepackage{wrapfig}
\usepackage{wrapfig}
\usepackage[most]{tcolorbox}
\usepackage{tabularx}
\tcbuselibrary{skins,raster}
\usepackage{titletoc}
\usepackage{enumitem}
\usepackage{placeins}

\definecolor{TC}{rgb}{0.56, 0.24, 0.20}
\definecolor{SC}{rgb}{0.12, 0.35, 0.43}

\definecolor{MedBlue}{HTML}{2F5D73}
\definecolor{MedBlueLight}{HTML}{F4F8FA}
\definecolor{MedBlueLine}{HTML}{B7C9D1}
\definecolor{MedText}{HTML}{1F3138}
\newcounter{noteHWctr} 
 
\newtcolorbox{paperboxdouble}[2][]{%
  float,
  floatplacement=t,
  width=\columnwidth,
  enhanced,
  colback=MedBlueLight,
  colframe=MedBlueLine,
  coltext=MedText,
  boxrule=0.7pt,
  arc=1.2mm,
  left=7pt,
  right=7pt,
  top=10pt,
  bottom=8pt,
  fonttitle=\bfseries,
  coltitle=white,
  colbacktitle=MedBlue,
  boxed title style={
    boxrule=0pt,
    arc=1mm,
    left=6pt,
    right=6pt,
    top=4pt,
    bottom=4pt
  },
  attach boxed title to top left={xshift=2mm,yshift*=-\tcboxedtitleheight/2},
  code={\refstepcounter{table}},
  title={Table~\thetable: #2},
  before skip=8pt,
  after skip=8pt,
  #1
}

\newtcblisting{promptboxfloat}[2][]{%
  enhanced,
  float,
  floatplacement=t,
  width=\columnwidth,
  colback=MedBlueLight,
  colframe=MedBlueLine,
  coltext=MedText,
  boxrule=0.5pt,
  arc=0.8mm,
  boxsep=1pt,
  left=4pt,
  right=4pt,
  top=4pt,
  bottom=4pt,
  fonttitle=\bfseries\small,
  coltitle=white,
  colbacktitle=MedBlue,
  boxed title style={
    boxrule=0pt,
    arc=0.8mm,
    left=4pt,
    right=4pt,
    top=2pt,
    bottom=2pt
  },
  attach boxed title to top left={xshift=2mm,yshift*=-\tcboxedtitleheight/2},
  title={#2},
  listing only,
  listing engine=listings,
  listing options={
    basicstyle=\ttfamily\scriptsize,
    breaklines=true,
    breakatwhitespace=false,
    columns=fullflexible,
    keepspaces=true,
    showstringspaces=false,
    upquote=true,
    aboveskip=0pt,
    belowskip=0pt,
    breakautoindent=false,
    breakindent=0pt,
    escapeinside={(*@}{@*)},
  },
  before skip=4pt,
  after skip=4pt,
  #1
}

\newtcblisting{promptbox}[2][]{%
  breakable,
  enhanced,
  width=\columnwidth,
  colback=MedBlueLight,
  colframe=MedBlueLine,
  coltext=MedText,
  boxrule=0.5pt,
  arc=0.8mm,
  boxsep=1pt,
  left=4pt,
  right=4pt,
  top=4pt,
  bottom=4pt,
  fonttitle=\bfseries\small,
  coltitle=white,
  colbacktitle=MedBlue,
  boxed title style={
    boxrule=0pt,
    arc=0.8mm,
    left=4pt,
    right=4pt,
    top=2pt,
    bottom=2pt
  },
  attach boxed title to top left={xshift=2mm,yshift*=-\tcboxedtitleheight/2},
  title={#2},
  listing only,
  listing engine=listings,
  listing options={
    basicstyle=\ttfamily\scriptsize,
    breaklines=true,
    breakatwhitespace=false,
    columns=fullflexible,
    keepspaces=true,
    showstringspaces=false,
    upquote=true,
    aboveskip=0pt,
    belowskip=0pt,
    breakautoindent=false,
    breakindent=0pt,
    escapeinside={(*@}{@*)},
  },
  before skip=4pt,
  after skip=4pt,
  #1
}

\usepackage{cleveref}
\usepackage{algorithm}
\usepackage{algpseudocode}

\title{\oursolution{}: Self-Play Reinforcement Learning from Case Reports for Medical Reasoning}

\author{%
  Hao~Wang\thanks{Equal contribution.}\textsuperscript{\phantom{$*$},1}\quad
  Zihan~Wang\footnotemark[1]\textsuperscript{\phantom{$*$},2}\quad
  Kai~Wu\footnotemark[1]\textsuperscript{\phantom{$*$},3}\quad
  Jian~Yao\thanks{Corresponding authors.}\textsuperscript{\phantom{$\dagger$},4}\\
  \textbf{Yiwen~Ye}\textsuperscript{5}\quad
  \textbf{Yongcan~Luo}\textsuperscript{1}\quad
  \textbf{Dapeng~Wu}\textsuperscript{1}\quad
  \textbf{Jiale~Chen}\textsuperscript{5}\quad
  \textbf{Haihua~Yang}\footnotemark[2]\textsuperscript{\phantom{$\dagger$},5}\\[0.5em]
  \textsuperscript{1}City University of Hong Kong\quad
  \textsuperscript{2}Northeastern University\\
  \textsuperscript{3}Tongji University\quad
  \textsuperscript{4}The Hong Kong Polytechnic University\quad
  \textsuperscript{5}ByteDance
}

\begin{document}

\maketitle

\begin{abstract}
Reinforcement learning (RL) is a promising way to improve the clinical reasoning abilities of large language models (LLMs), but current pipelines often rely on expert-labeled questions or costly synthetic-data generation. Meanwhile, the medical literature offers a wealth of high-quality unlabeled documents, including clinical case reports that trace the full clinical course from initial presentation to final treatment decisions. We study whether document-grounded self-play can convert these reports into effective RL training signals. We identify two challenges that limit existing self-play methods in this setting: \emph{knowledge-point fixation}, where the proposer repeatedly asks about a narrow set of salient clues in a source report, and \emph{difficulty-quality mismatch}, where appropriately difficult questions may still be poorly formed or shallow, limiting their value for reasoner training. To address these challenges, we propose \oursolution{}, a self-play framework with two key components: (a) a \emph{Knowledge-Conditioned Proposer}, which anchors question generation to case-specific knowledge points so that questions draw on a broader range of evidence from the source report; and (b) a \emph{Rubric-Judge Reasoner}, which augments answer-accuracy feedback with rubric-based quality feedback on clue integration, reasoning depth, and option quality, guiding the proposer toward reasoning-intensive questions that better support reasoner training. Extensive experiments across seven medically related benchmarks demonstrate that \oursolution{} outperforms existing self-play baselines and improves the SFT warm-start model by 3.1 points on average, highlighted by a 6.3-point gain on LiveClin. This approach offers an effective path toward continually improving medical reasoning ability through self-play RL. 
\end{abstract}

\section{Introduction}
Large language models (LLMs) have shown growing promise in medical reasoning, including differential diagnosis~\citep{m1,huatuo-1,meds3,medr3,agentic_diagnosis}, clinical decision support~\citep{virual_clinical,vrdoljak2025review,ong2025large,gaber2025evaluating}, and health question answering~\citep{health_bench,ehr-r1,yang2023one}. Many recent advances in medical LLMs have followed the common post-training paradigm of supervised fine-tuning (SFT) followed by reinforcement learning (RL), which has proven effective for improving reasoning ability. However, scaling this recipe remains costly as it depends on expert annotations and synthetic data pipelines, a bottleneck that becomes increasingly severe as models improve. In contrast, medicine already offers a continually growing collection of unlabeled documents (e.g., case reports) that contain rich real-world medical knowledge. This raises a fundamental question: \textit{can we continually improve medical reasoning by learning directly from unlabeled clinical corpora?}

A promising direction is self-play RL over text corpora~\citep{absolutezero,r-zero,DBLP:journals/corr/abs-2509-07414,DBLP:journals/corr/abs-2509-23863,spice,DARC,DBLP:journals/corr/abs-2601-18778,R-diverse,WIST}. In this paradigm, a \emph{Proposer} generates questions from either its internal knowledge or external documents, and a \emph{Reasoner} learns to solve them. The reasoner's performance provides feedback for updating the proposer, while the generated questions provide training data for improving the reasoner, forming an automated loop for deriving learning signals from unlabeled corpora. Although this paradigm has shown promising results in domains such as mathematics~\citep{DBLP:journals/corr/abs-2511-09907,DBLP:journals/corr/abs-2511-11881} and code~\citep{wei2025trainingsuperintelligentsoftwareagents}, directly applying it to the medical domain raises two challenges: (a) \emph{knowledge-point fixation}: medical case reports are information-dense and often contain multiple interdependent clinical clues, yet a naive proposer tends to anchor on the most salient evidence and repeatedly generate questions from the same narrow perspective; and (b) \emph{difficulty-quality mismatch}: existing self-play methods often use the reasoner's answer accuracy as the main feedback signal, treating intermediate-difficulty questions as useful training data. However, such questions may still be poorly constructed or require only shallow reasoning, making them poor candidates for improving medical reasoning through RL.

\begin{figure}[t]
  \centering
  \includegraphics[width=1.0\linewidth]{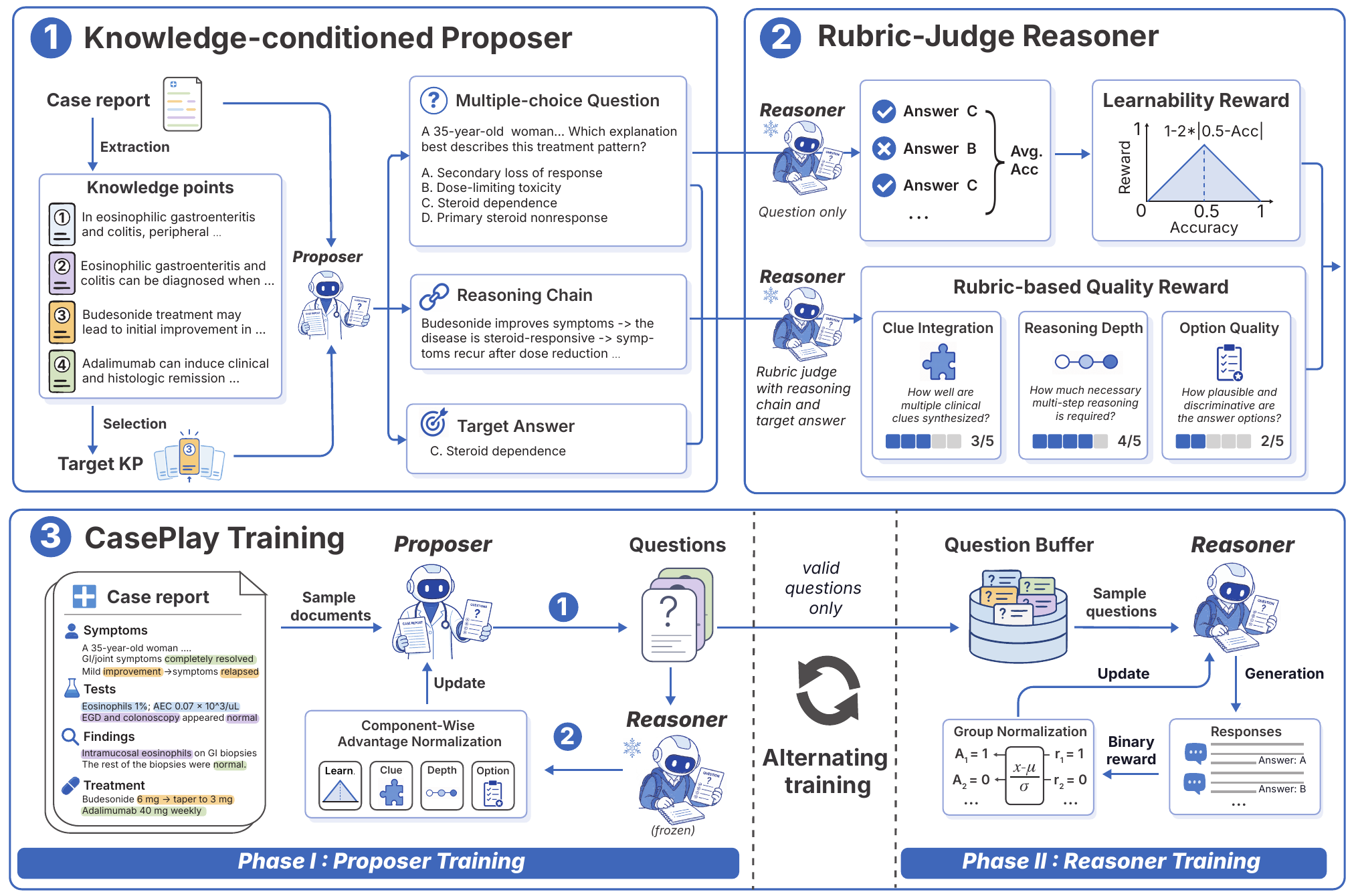}
  \caption{Overview of \oursolution{}.}
  \label{fig:overview}
\end{figure}

To address these challenges, we propose \textbf{\oursolution{}}, a self-play learning framework consisting of two key components: (a) a \textbf{Knowledge-Conditioned Proposer}, which conditions question generation on case-specific knowledge points to broaden coverage of each source report and reduce redundant questions around the same salient clue; and (b) a \textbf{Rubric-Judge Reasoner}, which supplements answer-accuracy feedback with rubric-based quality feedback along three dimensions: clue integration, reasoning depth, and option quality. This quality-aware feedback guides the proposer toward higher-quality questions that provide more useful training examples for the reasoner. Experiments on seven medically related benchmarks show that \oursolution{} improves the warm-start model by \textbf{3.1 points} on average and outperforms existing self-play baselines, with the largest gain on LiveClin~\citep{liveclin} (\textbf{+6.3 points}). Together, these results position \oursolution{} as a new way to leverage unlabeled case reports for improving medical reasoning through self-play RL. In summary, our contributions are threefold:

\begin{itemize}
\item We study self-play RL over clinical case reports for improving medical reasoning, and empirically identify two characteristic failure modes of existing methods: \emph{knowledge-point fixation} and \emph{difficulty-quality mismatch}.

\item We propose \oursolution{}, a medical self-play framework where a Knowledge-Conditioned Proposer uses case-specific knowledge points to encourage broader document coverage, while a Rubric-Judge Reasoner leverages rubric-based feedback as an additional quality signal for proposer optimization, helping generate less redundant and higher-quality training questions for improving medical reasoning.

\item We validate \oursolution{} through extensive experiments on 7 medically related benchmarks, where it improves the warm-start model by \emph{3.1 points} on average, outperforms medical reasoning models and self-play baselines, and remains effective across ablations, hyperparameter settings, and model scales.

\end{itemize}

\section{Pitfalls of Self-Play RL in Medical Reasoning}
\label{sec:pitfall}

\paragraph{Preliminaries.}
We formulate self-play training as an alternating game between a \emph{Proposer} $\pi_\theta^P$ and a \emph{Reasoner} $\pi_\phi^R$ over an unlabeled document corpus $\mathcal{D}$:
\begin{equation}
\mathcal{G}
=
\left\langle
\mathcal{M}^{P}_{1}(\phi_{0}),
\mathcal{M}^{R}_{1}(\theta_{1}),
\mathcal{M}^{P}_{2}(\phi_{1}),
\mathcal{M}^{R}_{2}(\theta_{2}),
\ldots,
\mathcal{M}^{P}_{T}(\phi_{T-1}),
\mathcal{M}^{R}_{T}(\theta_{T})
\right\rangle,
\label{eq:self_play_game}
\end{equation}
where $T$ is the number of self-play iterations. At iteration $t$, the proposer is first updated against the frozen reasoner $\pi_{\phi_{t-1}}^R$, and the reasoner is then updated using valid questions accumulated during the proposer-update phase. In the proposer phase $\mathcal{M}^{P}_{t}=(\mathcal{O}^{P},\mathcal{A}^{P},R^{P})$, the observation $o_t^P\in\mathcal{O}^{P}$ is a batch of clinical documents $\{d_{t,i}\}_{i=1}^{B}\subset\mathcal{D}$, the action $a_t^P\in\mathcal{A}^{P}$ is to generate grounded multiple-choice questions with target answers, and the reward $r_t^P$ is computed from the current reasoner's performance on these questions. In the reasoner phase $\mathcal{M}^{R}_{t}=(\mathcal{O}^{R},\mathcal{A}^{R},R^{R})$, the observation $o_t^R\in\mathcal{O}^{R}$ consists only of proposer-generated training questions, without access to the source documents; the action $a_t^R\in\mathcal{A}^{R}$ is to answer these questions; and the reward $r_t^R$ is a binary correctness reward.

However, existing self-play RL algorithms often yield limited gains when directly applied to medical corpora. To understand the limitations of current self-play methods in this setting, we analyze SPICE, a representative document-grounded baseline, and identify two recurring pitfalls in medical self-play: \emph{knowledge-point fixation} and \emph{difficulty-quality mismatch}. Details of the visualization are provided in Appendix~\ref{apdx:visual}.

\begin{figure}[!htbp]
  \centering
  \begin{subfigure}[t]{0.32\textwidth}
    \centering
    \includegraphics[width=\linewidth]{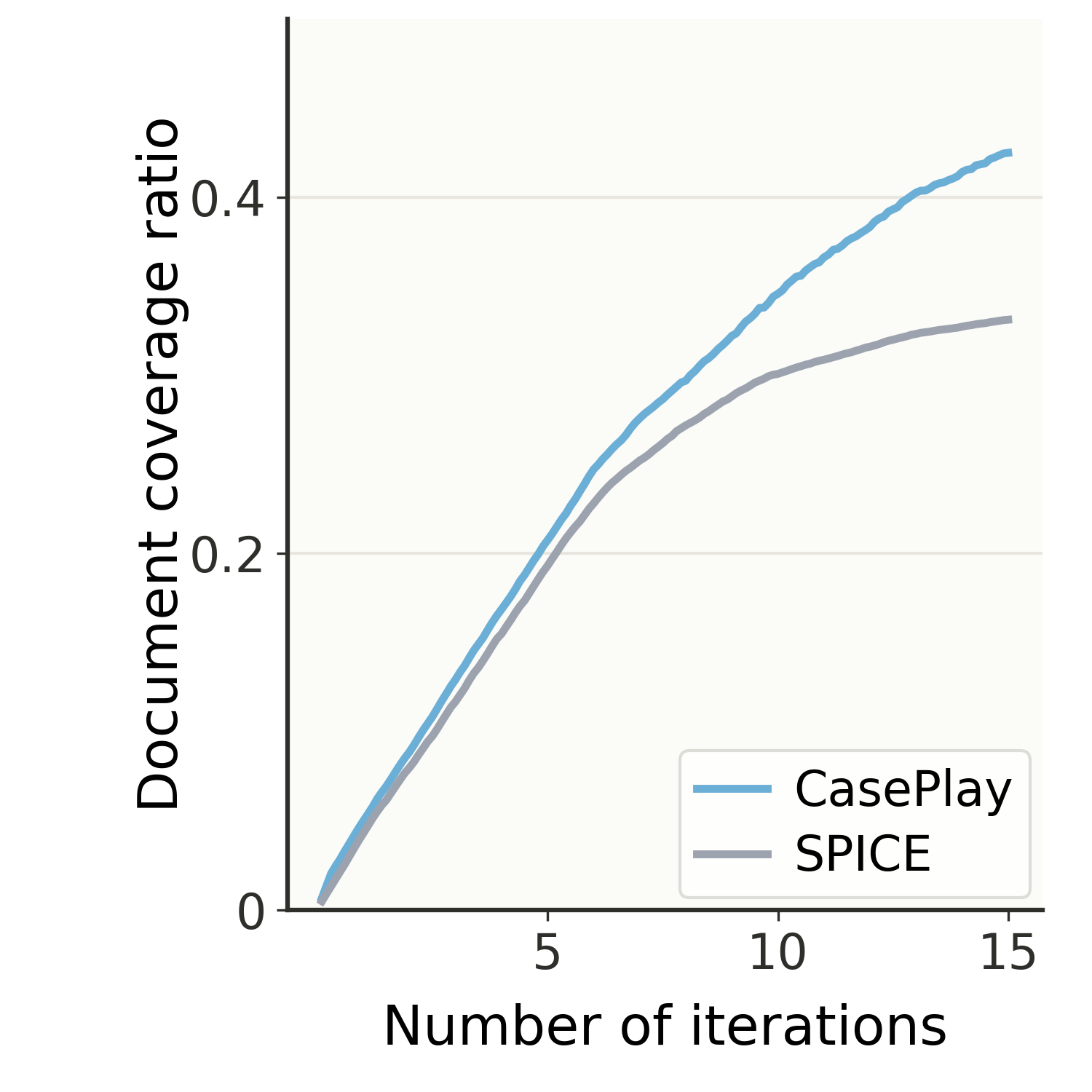}
    \caption{Source-document coverage.}
    \label{fig:pitfall_span_coverage}
  \end{subfigure}
  \hfill
  \begin{subfigure}[t]{0.32\textwidth}
    \centering
    \includegraphics[width=\linewidth]{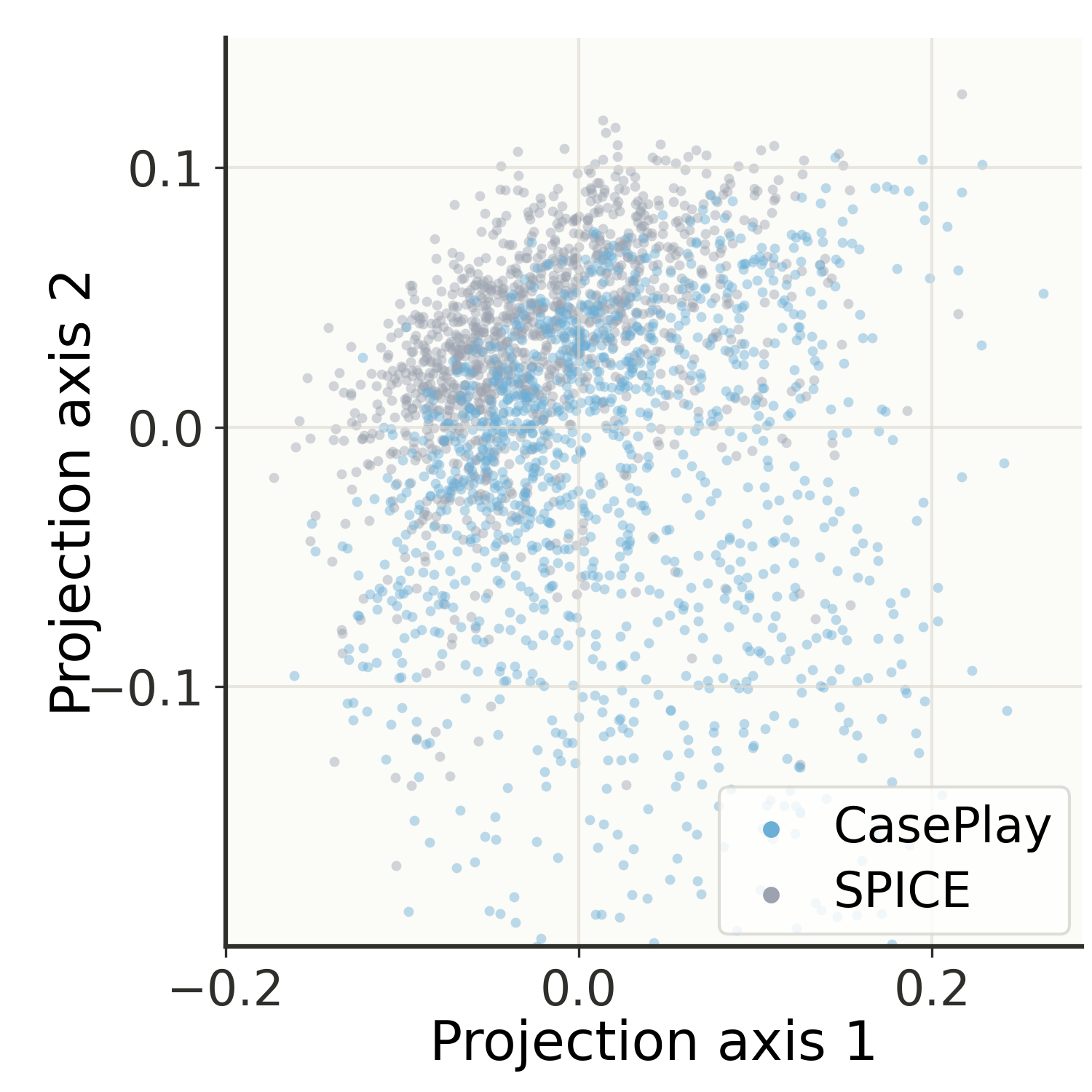}
    \caption{Question distribution.}
    \label{fig:pitfall_mds}
  \end{subfigure}
  \hfill
  \begin{subfigure}[t]{0.32\textwidth}
    \centering
    \includegraphics[width=\linewidth]{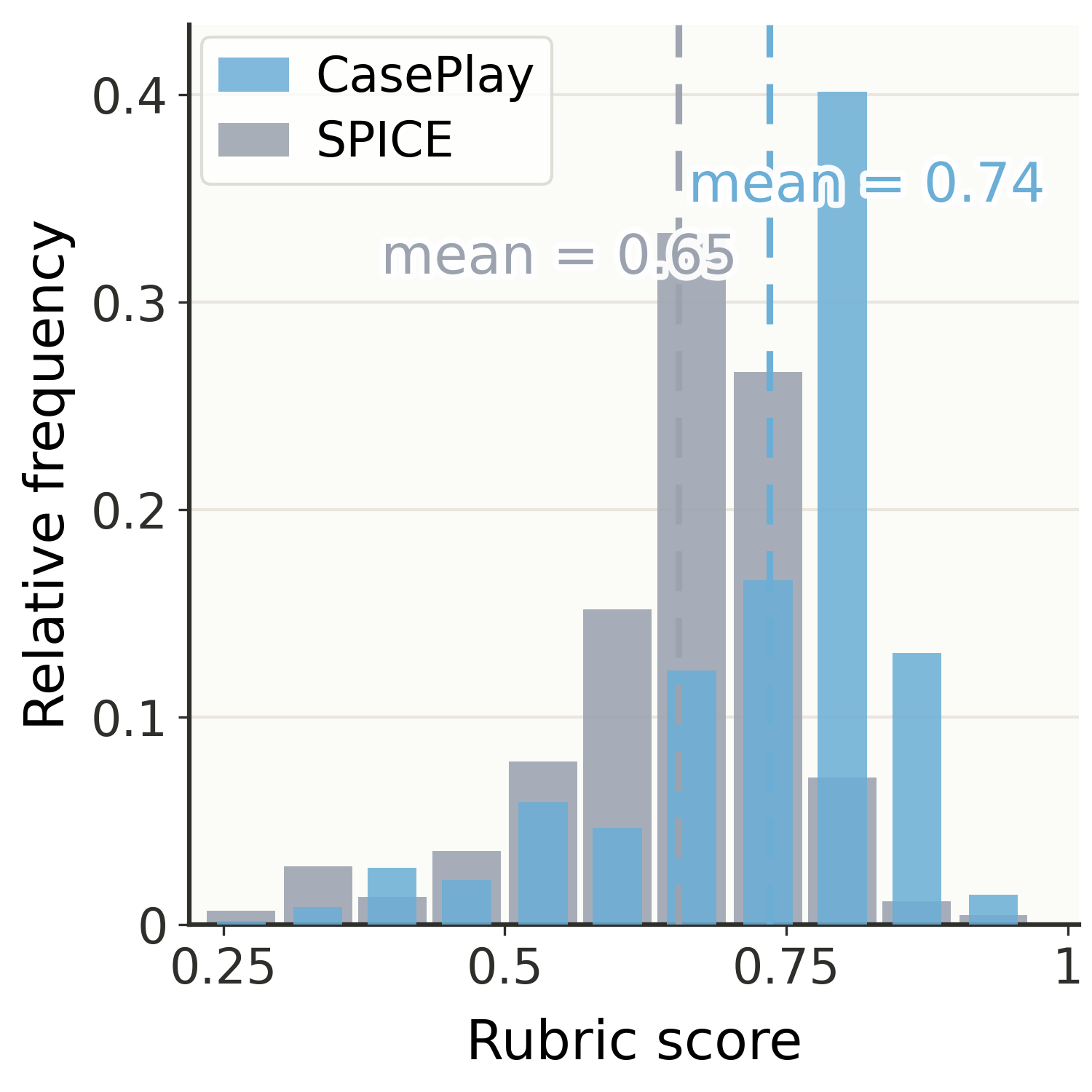}
    \caption{Quality at matched difficulty.}
    \label{fig:pitfall_rubric_dist}
  \end{subfigure}
\caption{(a) Cumulative source-document coverage of generated questions over training. (b) MDS~\citep{mds} projection of generated-question embeddings. (c) Rubric-based quality-score distribution among intermediate-difficulty questions with reasoner accuracy of 0.5.}
  \label{fig:pitfall_overview}
\end{figure}

\paragraph{Knowledge-Point Fixation.}
Clinical case reports are information-dense, often containing multiple clinical clues about symptoms, treatments, outcomes, and disease mechanisms. Ideally, document-grounded self-play should turn these diverse clues into diverse training questions. However, when the proposer is conditioned only on the full document, it tends to anchor on the most salient evidence and repeatedly generate questions from the same narrow subset of the report. Figure~\ref{fig:pitfall_overview}(a) shows that SPICE's source-document coverage quickly saturates at 0.33, compared with 0.42 for \oursolution{}, indicating that later rollouts mostly reuse already-covered content rather than explore new case evidence. The resulting questions are also less diverse: in Figure~\ref{fig:pitfall_overview}(b), SPICE forms a tighter cluster and occupies a smaller region than \oursolution{} in the projection space. We refer to this failure mode as \emph{knowledge-point fixation}, where an information-rich case report is reduced to redundant training questions from a narrow subset of clinical evidence. This observation indicates that the proposer needs an explicit mechanism to control which part of the case report each rollout should explore.

\begin{wraptable}[7]{r}{0.4\columnwidth}
\centering
\vspace{-1em}
\small
\captionsetup{font=small,skip=2pt,width=\linewidth}
\caption{Impact of question quality.}
\label{tab:difficulty_quality_mismatch}
\setlength{\tabcolsep}{3pt}
\resizebox{\linewidth}{!}{%
\begin{tabular}{@{}lcc@{}}
\toprule
Method & Avg. Acc. & Resp. Len. \\
\midrule
Warm-start Model & 63.1 & 3794.5 \\
\quad + RL (Random) & 63.5 & 3025.4 \\
\quad + RL (Filtered) & \textbf{64.3} & \textbf{4932.8} \\
\bottomrule
\end{tabular}}
\vspace{-1em}
\end{wraptable}

\paragraph{Difficulty-Quality Mismatch.}
Existing self-play methods often update the proposer using reasoner-estimated difficulty, favoring questions that are neither too easy nor too hard~\citep{absolutezero,r-zero,spice}. However, intermediate difficulty does not guarantee training value in medical reasoning: such questions may still be poorly constructed, or require only shallow reasoning, making them poor candidates for improving medical reasoning through RL. To examine this issue, we select questions with empirical reasoner accuracy of 0.5 and evaluate their quality using the rubrics introduced in Section~\ref{sec:solution:rubric}. As shown in Figure~\ref{fig:pitfall_overview}(c), even within this intermediate-difficulty set, SPICE produces many low-quality questions, with 31.3\% receiving rubric scores no higher than 0.6. We then construct two matched training sets from these questions: a random subset and a rubric-filtered subset that keeps questions with average rubric score above 0.6. As shown in Table~\ref{tab:difficulty_quality_mismatch}, RL training on the rubric-filtered set improves average benchmark accuracy from 63.1 to 64.3, whereas random selection reaches only 63.5 and reduces the average response length. This highlights the need for quality-aware feedback that goes beyond difficulty and favors well-formed, reasoning-intensive questions.

\section{Proposed Solution: \oursolution{}}
In this paper, we propose \textbf{\oursolution{}}, a self-play framework for improving medical reasoning from case reports. The central idea of \oursolution{} is to turn two implicit choices in document-grounded self-play into explicit controls: \emph{which clinical focus} in a case report the proposer should explore, and \emph{how} to evaluate the quality of generated questions for training. \oursolution{} implements these controls with a \emph{Knowledge-Conditioned Proposer}, which guides question generation with case-specific knowledge points, and a \emph{Rubric-Judge Reasoner}, which provides quality-aware feedback for proposer learning.

\subsection{Knowledge-Conditioned Proposer}
Prior document-grounded self-play methods typically condition the proposer on the entire document~\citep{spice,DARC}, leaving the clinical focus implicit during the generation process. Conceptually, generating $o$ from a case report $d$ can be viewed as marginalizing over a latent focus $z$:
\begin{equation}
\pi_\theta(o \mid d)
=
\sum\nolimits_{z}
\pi_\theta(z \mid d)\pi_\theta(o \mid d,z),
\label{eq:latent_focus_factorization}
\end{equation}
where $o$ denotes the proposer output. When $z$ is left implicit, the proposer must choose the focus while writing the question, often biasing generation toward salient clues such as the final diagnosis. Chunking offers a crude focus proxy, but can remove the temporal or treatment context needed for faithful medical questions. We therefore make $z$ explicit as a document-grounded knowledge point: a discrete clinical focus in the report that guides what the proposer should ask about, while the full report remains available as context.

\paragraph{Knowledge-Point Extraction.}
Before self-play, for each document $d \in \mathcal{D}$, we use the proposer to extract candidate knowledge points from the source report, remove redundant or overlapping candidates, and retain at most $M$ of them, denoted as $K(d)=\{k_1,\dots,k_{M}\}$. Each knowledge point $k_m$ summarizes a clinically meaningful aspect of the case, such as a symptom progression, diagnostic clue, or outcome pattern, and is supported by explicit evidence in the source document, as illustrated in Appendix Figure~\ref{fig:caseplay_case_study}. We keep $K(d)$ fixed throughout self-play to provide stable clinical focuses and avoid introducing additional non-stationarity into proposer optimization. Details of the extraction procedure are provided in Appendix~\ref{apdx:detail:kg_extraction}.

\paragraph{Knowledge-Conditioned Question Generation.}
\oursolution{} implements the latent-focus view in Eq.~(\ref{eq:latent_focus_factorization}) by factorizing each proposer rollout into knowledge-point selection and knowledge-conditioned generation. Specifically, for the $j$-th rollout of document $d$:
\begin{equation}
\tilde{\pi}_\theta^P(k,o \mid d,K(d),j)
=
\pi^{\mathrm{sel}}(k \mid K(d),j)\,
\pi_\theta^P(o \mid d,k),
\label{eq:kp_conditioned_generation}
\end{equation}
where $k$ is the selected knowledge point and $o=(q,c,y)$ consists of the generated multiple-choice question $q$, proposer-provided reasoning chain $c$, and target answer $y$. In our implementation, $\pi^{\mathrm{sel}}$ is a fixed round-robin selector, which avoids introducing an additional learned selector while remaining computationally efficient and empirically effective. Specifically, for $K(d)=\{k_1,\dots,k_M\}$:
\begin{equation}
\pi^{\mathrm{sel}}(k = k_m \mid K(d),j)
=
\mathbb{I}\!\left[m = 1 + ((j-1)\bmod M)\right].
\label{eq:round_robin_selector}
\end{equation}
Thus, repeated rollouts of the same case report are anchored to different knowledge points, encouraging the proposer to cover multiple clinical focuses rather than repeatedly asking about the same salient one. We compare alternative knowledge-point selection strategies in Appendix~\ref{apdx:exp:kg_selection}. After generation, each output $o$ is evaluated by the current reasoner to produce feedback rewards for the proposer update, as described next.

\subsection{Rubric-Based Feedback from the Reasoner}
\label{sec:solution:rubric}
As discussed in Section~\ref{sec:pitfall}, answer accuracy mainly reflects a question's difficulty for the current reasoner, but does not necessarily capture whether the question is useful for RL training. Therefore, \oursolution{} evaluates the generated question with two reward components: \emph{learnability}, estimated from the reasoner's answer accuracy without access to the source report, and \emph{quality}, judged by rubric-based criteria using the target answer and proposer-provided reasoning chain.

\paragraph{Learnability Reward.}
Given a proposer-generated multiple-choice question $q$ with target answer $y$, the reasoner samples $N_{\mathrm{acc}}$ answers to $q$ without access to the source case report, yielding an estimated accuracy $\mathrm{acc}(q)$. Following~\citep{r-zero}, the learnability reward is defined as:
\begin{equation}
r^{\mathrm{learn}}
=
1 - 2\left|\mathrm{acc}(q)-0.5\right|,
\label{eq:learnability_reward}
\end{equation}
which is maximized when $\mathrm{acc}(q)=0.5$, encouraging the proposer to generate questions that are neither too easy nor too hard for the current reasoner.

\paragraph{Rubric-Based Quality Reward.}
Medical questions with similar difficulty can differ substantially in the reasoning they elicit. To assess question quality, we ask the reasoner to judge the generated question together with the target answer $y$ and proposer-provided reasoning chain $c$. With the proposer-provided answer and reasoning chain, the reasoner can more directly assess whether the intended reasoning path is coherent and whether the options are discriminative, without inferring the rationale from the question alone. Formally, the quality reward is calculated as:
\begin{equation}
\mathbf{r}^{\mathrm{qual}}
=
\left(
r^{\mathrm{clue}},
r^{\mathrm{depth}},
r^{\mathrm{option}}
\right)
=
\pi^{J}_\phi(o),
\label{eq:quality_reward}
\end{equation}
where the three dimensions capture complementary aspects of question utility for RL training. \emph{Clue integration} measures whether answering the question requires synthesizing multiple clinical clues; \emph{reasoning depth} measures whether the question requires multi-hop medical reasoning rather than surface-level fact matching; and \emph{option quality} measures whether the answer options include plausible and discriminative distractors. Detailed rubric prompts are provided in Appendix~\ref{apdx:detail:rubrics}.

\subsection{Alternating Self-Play Optimization}

\oursolution{} is trained over multiple self-play iterations. Each iteration contains two phases: a \emph{proposer-update phase} and a \emph{reasoner-update phase}. In each phase, we update one model for multiple steps while keeping the other frozen. Both models are optimized with the GRPO~\citep{grpo} objective:
\begin{equation}
\mathcal{L}_{\mathrm{GRPO}}
=
-\mathbb{E}_{i}\left[
\min\left(
\rho_i \hat{A}_i,\;
\mathrm{clip}(\rho_i,1-\varepsilon,1+\varepsilon)\hat{A}_i
\right)
\right],
\label{eq:grpo_obj}
\end{equation}
where $\rho_i=\frac{\pi(o_i\mid x_i)}{\pi_{\mathrm{old}}(o_i\mid x_i)}$ is the importance-sampling ratio for the corresponding phase, with $x_i$ denoting the model input and $o_i$ denoting the sampled output. The advantage $\hat{A}_i$ is normalized within each sampled group, with phase-specific reward definitions described below.

\paragraph{Proposer Update with Component-Wise Advantage Normalization.}
In the proposer-update phase, we sample a batch of documents, and for each document the proposer generates a group of $N_p$ candidate outputs according to the knowledge-conditioned rollout in Eq.~(\ref{eq:kp_conditioned_generation}). The frozen reasoner evaluates each valid candidate with the learnability reward in Eq.~(\ref{eq:learnability_reward}) and the rubric-based quality rewards in Eq.~(\ref{eq:quality_reward}). If a candidate violates the required format, we assign the format penalty $r^{\mathrm{format}}$ to all reward components and skip reasoner evaluation. Since these reward components capture different aspects of question utility, we normalize each component separately within the candidate group before aggregation. For a generated question $i$, the final proposer advantage $\hat{A}_i^P$ is computed as:
\begin{equation}
\tilde{r}_i^s
=
\frac{r_i^s-\mathrm{mean}_{j}(r_j^s)}
{\mathrm{std}_{j}(r_j^s)+\epsilon},
\qquad
\hat{A}_i^P
=
\tilde{r}_i^{\mathrm{learn}}
+
\lambda
\left(
\tilde{r}_i^{\mathrm{clue}}
+
\tilde{r}_i^{\mathrm{depth}}
+
\tilde{r}_i^{\mathrm{option}}
\right),
\label{eq:component_advantage}
\end{equation}
where $s\in\{\mathrm{learn},\mathrm{clue},\mathrm{depth},\mathrm{option}\}$ indexes the reward component and $\lambda$ controls the overall contribution of rubric-based feedback.

\paragraph{Reasoner Update.}
During the proposer-update phase, generated valid questions are stored in a training buffer $\mathcal{B}$. After the proposer phase finishes, we sample a subset of questions from this buffer to update the reasoner. For each sampled question, the reasoner generates $N_r$ responses without access to the source report, and each response receives a binary reward: $1$ if the final answer matches the target answer and $0$ otherwise. These rewards are normalized within the response group to obtain $\hat{A}_i^R$, and the reasoner is updated by Eq.~(\ref{eq:grpo_obj}). The pseudocode of \oursolution{} is provided in Appendix~\ref{apdx:detail:training_procedure}.

\section{Experiments}
\subsection{Setup}

\paragraph{Models and Baselines.}
We evaluate \oursolution{} on two model families and multiple scales, including Qwen3-4B-Base~\citep{qwen3}, Qwen3-8B-Base~\citep{qwen3}, and OctoThinker-3B-Hybrid-Base~\citep{wang2025octothinker}. We compare against two categories of baselines. The first category consists of recent medical reasoning models, including HuatuoGPT-o1-8B~\citep{huatuo-1}, AlphaMed-8B~\citep{alphamed}, and Fleming-R1-7B~\citep{fleming-r1}. The second category consists of self-play methods, including R-Zero~\citep{r-zero}, SPICE~\citep{spice}, and DARC~\citep{DARC}. Details of baseline implementations are provided in Appendix~\ref{apdx:exp:baselines}.

\paragraph{Training Data.}
For training data collection, we collect open-access case reports from PubMed Central (PMC~\citep{pmc}) published between 2020 and 2025. We apply a strict filtering pipeline to remove incomplete reports, image-dominant articles, and documents with potential benchmark leakage, yielding a final corpus of 4,281 case reports. For each filtered document, we extract up to $M=4$ knowledge points to support knowledge-conditioned question generation. To ensure stable instruction following and output formatting, we construct a small SFT warm-start dataset from the training case reports. Both \oursolution{}~and all self-play baselines use this dataset for warm-start initialization to ensure a fair comparison. Additional details of training data collection are provided in Appendix~\ref{apdx:data}.

\paragraph{Evaluation.}
We evaluate \oursolution{} on seven benchmarks across three categories. First, we test whether our method improves performance on expert-level medical reasoning benchmarks, including MedXpert-Text~\citep{medXpertQA}, MedBullets-OP4/OP5~\citep{op5}, and MedQA-USMLE~\citep{medqa-usmle}. Second, we evaluate clinical case reasoning with LiveClin~\citep{liveclin}, which focuses on patient narratives and requires integrating case-specific clues from clinical context. Third, we evaluate transfer to scientific QA using MMLU-Pro~\citep{mmlu_pro} (Health and Biology subsets) and GPQA~\citep{gpqa} (Genetics and Molecular Biology subset). For each benchmark question, we sample 8 responses and report the average accuracy as the final metric. Detailed benchmark descriptions are provided in Appendix~\ref{apdx:exp:benchmarks}.

\paragraph{Implementation Details.}
We train \oursolution{} for 15 iterations, with each iteration consisting of 10 proposer-update steps and 10 reasoner-update steps. The rubric weight $\lambda$ is set to 0.2. For both the proposer and the reasoner, we use a training batch size of 128 and a learning rate of $1\times 10^{-6}$. During training rollouts, we use temperature 1.0 and top-$p$ 0.95, with a maximum input length of 4,096 tokens and a maximum output length of 8,192 tokens. During evaluation, we set the sampling temperature to 0.6 and top-$p$ to 0.95, while keeping the other generation settings unchanged. Full experimental hyperparameters are listed in Appendix~\ref{apdx:exp:training}.

\definecolor{AvgHighlight}{HTML}{EAF3FB}
\definecolor{GainGreen}{HTML}{2F6F4E}
\begin{table*}[!htbp]
\centering
\caption{Main results on seven medically related benchmarks.}
\label{tab:main_results}
\setlength{\tabcolsep}{2.0pt}
\renewcommand{\arraystretch}{1.18}
\resizebox{0.95\textwidth}{!}{%
\begin{tabular}{p{0.23\textwidth}ccccccc>{\columncolor{AvgHighlight}}c}
\toprule
\multirow{2}{*}{\textbf{Method}}
& \multicolumn{4}{c}{\textbf{Expert Medical Reasoning}}
& \multicolumn{1}{c}{\textbf{Clin. Case}}
& \multicolumn{2}{c}{\textbf{Science QA}}
& \cellcolor{white} \\
\cmidrule(lr){2-5} \cmidrule(lr){6-6} \cmidrule(lr){7-8}
& MedXpert
& MB-OP4
& MB-OP5
& MedQA
& LiveClin
& MMLU.
& GPQA
& Avg. \\
\midrule
\multicolumn{9}{c}{\textbf{Base Models}} \\
Qwen3-8B-Base~\citep{qwen3}
& 12.3 & 41.2 & 35.3 & 50.8 & 48.6 & 56.4 & 42.2 & 41.0 \\
\quad + SFT (Warm-Start)
& 22.7 & 72.9 & 68.1 & 83.6 & 66.0 & 75.3 & 53.4 & 63.1 \\
\midrule
\multicolumn{9}{c}{\textbf{Medical Reasoning Models}} \\
HuatuoGPT-o1-8B~\citep{huatuo-1}
& 15.4 & 52.6 & 45.2 & 72.6 & 55.4 & 63.1 & 50.4 & 50.7 \\
AlphaMed-8B~\citep{alphamed}
& 21.4 & 62.4 & 58.7 & 76.2 & 62.5 & 66.7 & 52.1 & 57.1 \\
Fleming-R1-7B~\citep{fleming-r1}
& 19.1 & 57.8 & 54.1 & 75.1 & 60.2 & 69.3 & 48.7 & 54.9 \\
\midrule
\multicolumn{9}{c}{\textbf{Self-Play RL Methods}} \\
R-Zero~\citep{r-zero}
& 19.7 & 72.9 & 68.6 & 83.5 & 66.0 & 73.3 & 52.9 & 62.4 \\
SPICE~\citep{spice}
& 21.8 & 74.2 & 70.0 & 83.2 & 67.1 & 75.6 & 54.2 & 63.7 \\
DARC~\citep{DARC}
& 23.5 & 75.7 & 69.8 & 84.8 & 70.5 & 75.3 & 55.2 & 65.0 \\
\midrule
\multicolumn{9}{c}{\textbf{Proposed Solution}} \\
\textbf{\oursolution{}}
& \textbf{25.1} & \textbf{76.9} & \textbf{70.1} & \textbf{85.4}
& \textbf{72.3}
& \textbf{75.9} & \textbf{57.6}
& \textbf{66.2} \\
\textit{Impr. vs SFT Model}
& \textcolor{GainGreen}{\textbf{+2.4}}
& \textcolor{GainGreen}{\textbf{+4.0}}
& \textcolor{GainGreen}{\textbf{+2.0}}
& \textcolor{GainGreen}{\textbf{+1.8}}
& \textcolor{GainGreen}{\textbf{+6.3}}
& \textcolor{GainGreen}{\textbf{+0.6}}
& \textcolor{GainGreen}{\textbf{+4.2}}
& \textcolor{GainGreen}{\textbf{+3.1}} \\
\midrule
\multicolumn{9}{c}{\textbf{Ablation Study}} \\
\,~w/o knowledge-point
& 23.3 & 75.8 & 69.4 & 84.7 & 69.8 & 75.2 & 56.3 & 64.9 \\
\,~w/o rubric judge
& 22.5 & 75.0 & 68.4 & 83.5 & 70.8 & 74.5 & 55.2 & 64.3 \\
\,~w/o proposer update
& 21.4 & 74.1 & 67.9 & 83.2 & 68.4 & 73.8 & 53.9 & 63.2 \\
\bottomrule
\end{tabular}%
}
\end{table*}

\subsection{Performance Comparison}
As shown in Table~\ref{tab:main_results}, \oursolution{} achieves the best performance on all seven benchmarks. Compared with the SFT warm-start model, \oursolution{} improves the average score from 63.1\% to 66.2\%, yielding a +3.1 point gain. The largest gain appears on LiveClin~\citep{liveclin} (+6.3 points), because its patient-case narratives require integrating symptoms, diagnostic clues, and treatment context, which matches the reasoning ability improved by self-play over clinical case reports. Among self-play baselines, R-Zero does not use external documents during question generation and slightly underperforms the warm-start model on average. Compared with document-grounded baselines such as SPICE and DARC, \oursolution{} achieves stronger gains, indicating that medical self-play benefits from broader coverage of case-specific knowledge points and quality-aware proposer feedback.

\subsection{Ablation Study}
\label{sec:ablation}
\begin{figure*}[!htbp]
  \centering
  \begin{subfigure}[t]{0.48\textwidth}
    \centering
    \includegraphics[width=\linewidth]{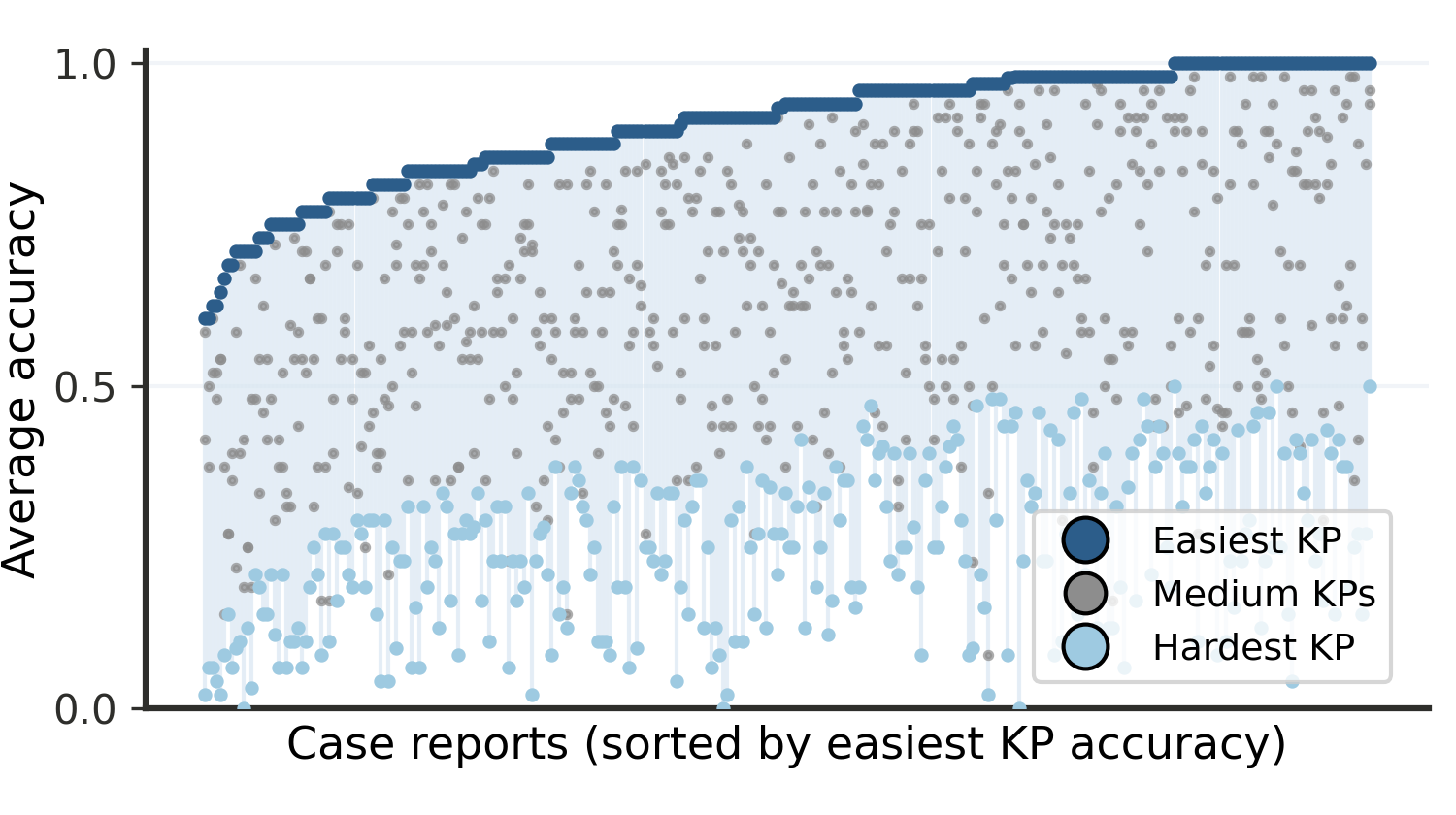}
    \caption{Within-case difficulty variation.}
    \label{fig:within_case_kp_variation}
  \end{subfigure}
  \hfill
  \begin{subfigure}[t]{0.48\textwidth}
    \centering
    \includegraphics[width=\linewidth]{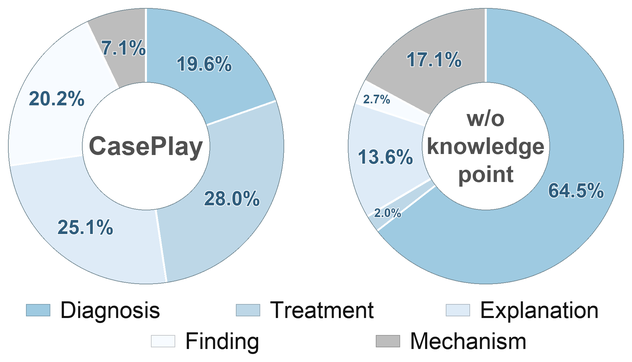}
    \caption{Question-type distribution.}
    \label{fig:question_distribution}
  \end{subfigure}
  \caption{(a) Different knowledge points from the same report yield substantially different difficulty. (b) Knowledge-conditioned generation produces a more balanced distribution of question types.}
  \label{fig:kp_diversity_analysis}
  \vspace{-2mm}
\end{figure*}

\paragraph{Knowledge-Conditioned Generation Diversifies Within-Case Questions.}
As shown in Figure~\ref{fig:kp_diversity_analysis}(a), we randomly sample 10\% of the training documents and visualize the reasoner's accuracy on questions generated from different knowledge points within the same case report. Questions grounded in different knowledge points often have substantially different difficulty, indicating that they probe distinct aspects of the case and provide complementary learning opportunities. We then remove knowledge-point conditioning during question generation and compare the resulting question-type distribution in Figure~\ref{fig:kp_diversity_analysis}(b). Without knowledge-point conditioning, generation becomes highly skewed toward diagnosis-related questions, which account for 64.5\% of all generated questions. In contrast, \oursolution{} produces a more balanced distribution across the five clinical question types. Consistent with this improved diversity, \oursolution{} outperforms the variant without knowledge points by 1.3 point on average, improving from 64.9 to 66.2 as shown in Table~\ref{tab:main_results}.

\paragraph{Necessity of Proposer Updates.}
To verify the benefit of proposer updates over static question generation, we conduct an additional ablation in which the proposer is frozen at the SFT warm-start checkpoint throughout training, while the reasoner is still updated on the generated questions. As shown in the last row of Table~\ref{tab:main_results}, this variant reaches only 63.2 average accuracy, 3.0 points lower than \oursolution{}, and also underperforms other document-grounded self-play methods. This result highlights the necessity of proposer adaptation, where the proposer adjusts its questions according to feedback from the current reasoner to keep producing useful training data.

\paragraph{Rubric Judge Improves Question Quality.}
\begin{wrapfigure}{r}{0.4\columnwidth}
\vspace{-5mm}
\centering
\captionsetup{font=small,skip=2pt}
\includegraphics[width=\linewidth]{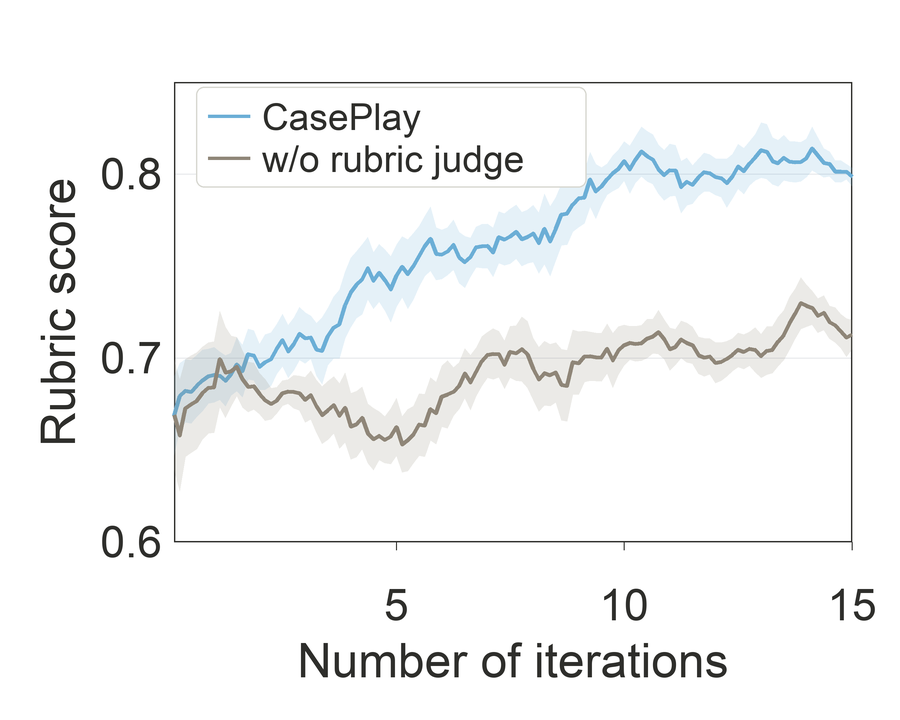}
\vspace{-3mm}
\caption{Training curve of rubric score.}
\label{fig:rubric_average}
\vspace{-1em}
\end{wrapfigure}
We compare the average rubric scores of questions generated by \oursolution{} and by the variant without rubric-judge feedback in Figure~\ref{fig:rubric_average}. For the latter, we still score generated questions with the reasoner for analysis, but exclude rubric rewards from proposer updates. Without rubric-judge feedback, the rubric score remains lower and less stable, indicating that learnability-based feedback alone is insufficient to promote high-quality medical question generation. In contrast, \oursolution{} steadily increases the average rubric score, helping the proposer generate questions with better clue integration, greater reasoning depth, and more discriminative options. This improvement in question quality translates into better reasoner performance: as shown in Table~\ref{tab:main_results}, \oursolution{} improves the average score from 64.3 to 66.2 compared with the variant without rubric judge.

\paragraph{Reliability of Reasoner-Based Rubric Judging.}
\begin{wraptable}{r}{0.40\columnwidth}
\vspace{-0.8em}
\centering
\small
\captionsetup{font=small,skip=2pt}
\caption{Agreement with DeepSeek-V3.2.}
\label{tab:rubric_corr}
\setlength{\tabcolsep}{5pt}
\begin{tabular}{@{}lc@{}}
\toprule
Rubric Dimension & Pearson Corr. $\uparrow$ \\
\midrule
Clue Integration    & 0.783 \\
Reasoning Depth     & 0.752 \\
Option Quality      & 0.627 \\
Average Score             & 0.705 \\
\bottomrule
\end{tabular}
\vspace{-1.0em}
\end{wraptable}
Since the rubric scores are produced by the current reasoner rather than a stronger external judge, an important concern is whether these scores provide reliable quality feedback. To assess this, we compare the reasoner's rubric scores with scores assigned by DeepSeek-V3.2~\citep{dpsk-v32} using the same rubric prompt. As shown in Table~\ref{tab:rubric_corr}, the reasoner achieves strong agreement with DeepSeek-V3.2 on \emph{clue integration} and \emph{reasoning depth}, with Pearson correlations of 0.783 and 0.752, respectively. The correlation is lower on \emph{option quality}, likely because judging distractor discriminativeness requires broader medical knowledge. The correlation for the average rubric score reaches 0.705, indicating that the reasoner provides a sufficiently reliable quality signal for proposer optimization.

\subsection{Hyperparameter Sensitivity}
\paragraph{Impact of the Number of Knowledge Points.}

\begin{figure*}[!htbp]
  \centering
  \begin{subfigure}[t]{0.48\textwidth}
    \centering
    \includegraphics[width=\linewidth]{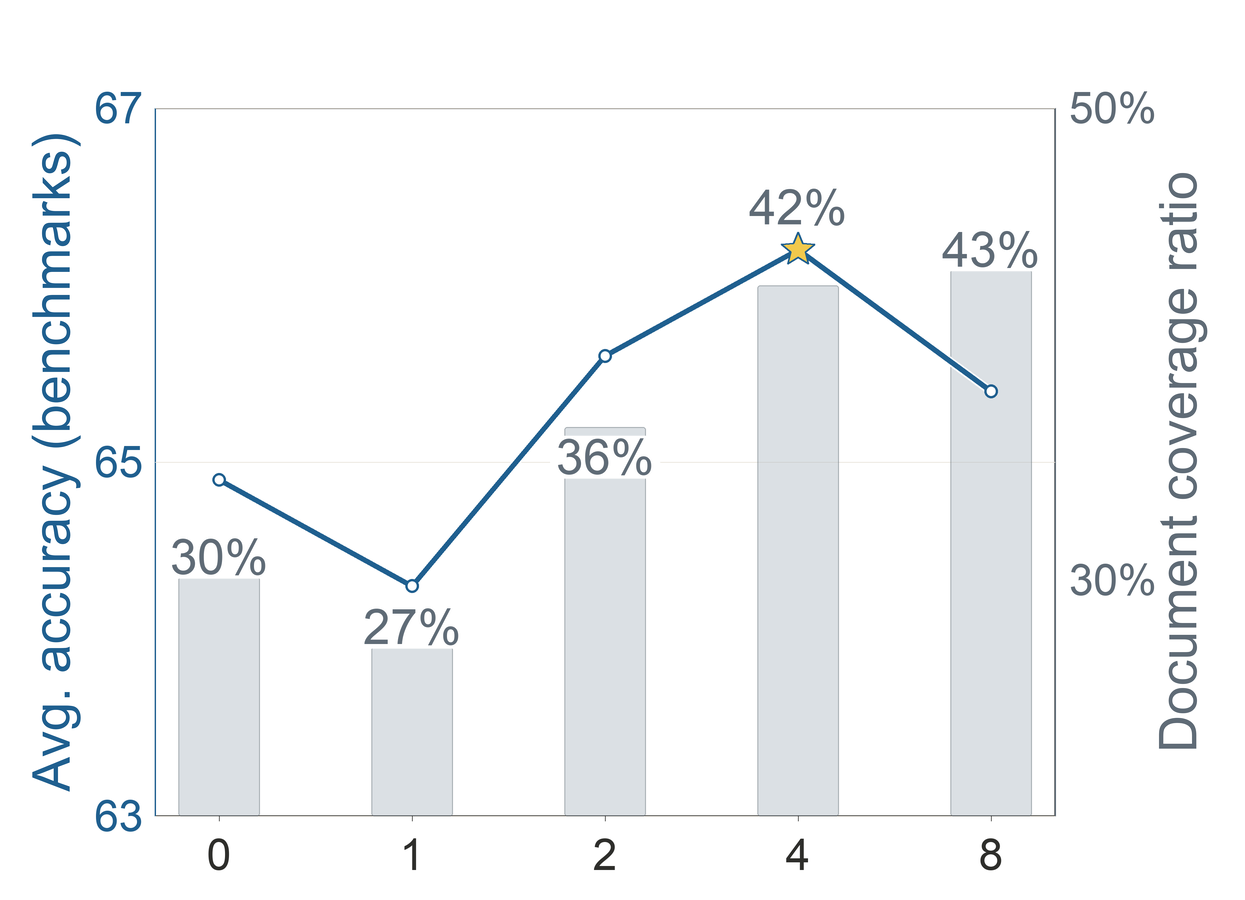}
    \caption{Number of knowledge points $M$}
    \label{fig:kp_count_analysis}
  \end{subfigure}
  \hfill
  \begin{subfigure}[t]{0.48\textwidth}
    \centering
    \includegraphics[width=\linewidth]{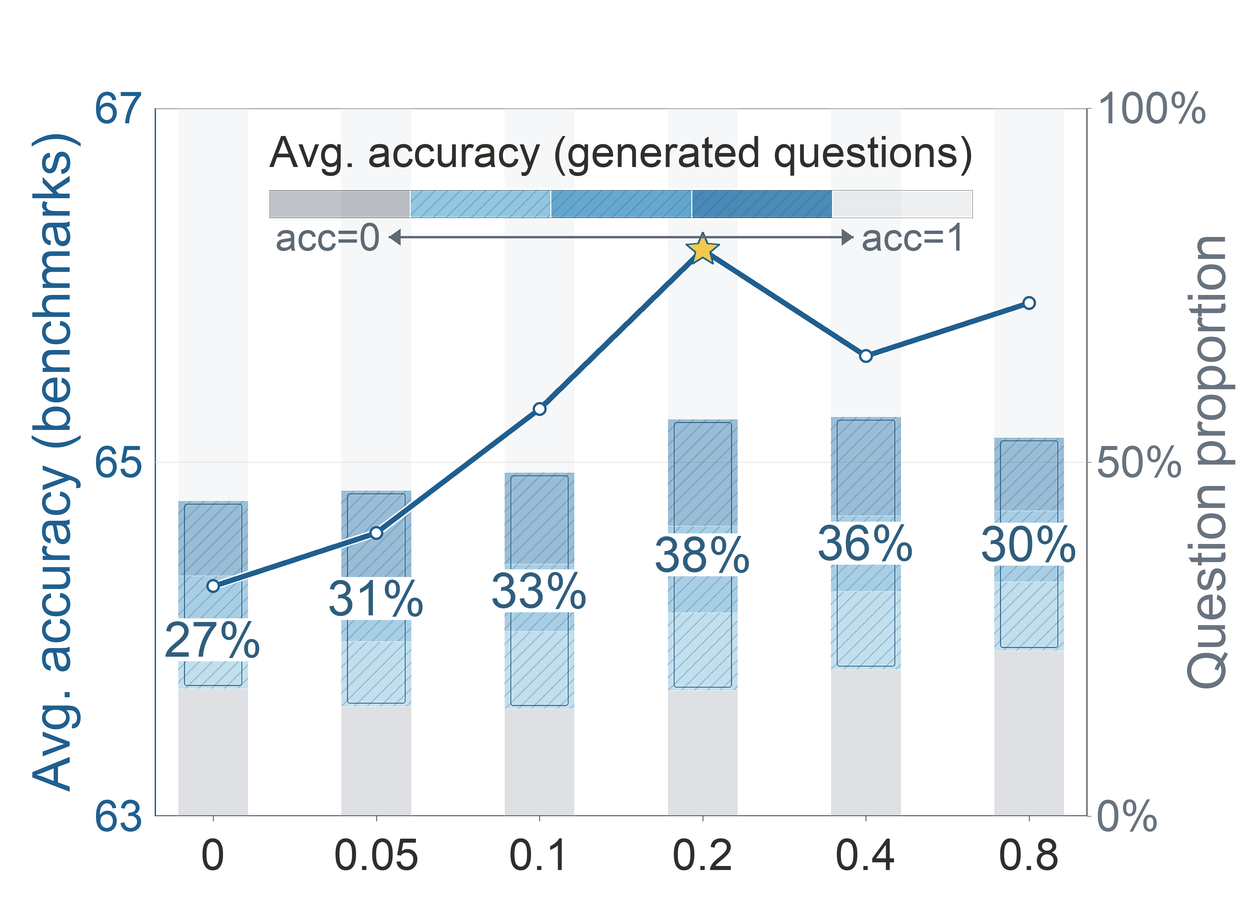}
    \caption{Rubric weight $\lambda$}
    \label{fig:rubric_weight_analysis}
  \end{subfigure}
  \caption{Sensitivity to (a) the number of knowledge points $M$ and (b) the rubric weight $\lambda$.}
  \label{fig:hyperparam_sensitivity}
\end{figure*}
The number of extracted knowledge points controls the granularity of clinical focus during question generation. We vary the maximum number of knowledge points per document, including the no-knowledge-point setting, and report the results in Figure~\ref{fig:hyperparam_sensitivity}(a). 
With too few knowledge points, the proposer conditions generation on a limited set of clinical focuses, leading to lower document coverage and weaker performance. As $M$ increases, document coverage ratio improves, rising from 27\% at $M=1$ to 43\% at $M=8$. However, when $M$ becomes larger than 4, document coverage improves only marginally while downstream accuracy begins to decline, indicating that too many knowledge points can make the clinical focuses overly fine-grained or weakly distinguishable, weakening the quality of generated questions. We therefore set $M=4$ as the default number of knowledge points per case report.

\paragraph{Impact of Rubric Weight.}
The rubric weight controls the trade-off between learnability and question quality during proposer optimization. In Figure~\ref{fig:hyperparam_sensitivity}(b), we report the proportion of learnable questions in GRPO (i.e., $0<\mathrm{acc}(q)<1$). When $\lambda$ is small, the proposer is primarily guided by estimated difficulty, so it may generate questions that are non-trivial for the reasoner but still shallow or poorly constructed. Increasing $\lambda$ to 0.2 raises the learnable-question proportion from 27\% to 38\% and yields the best accuracy, indicating that rubric feedback helps align difficulty with clinically meaningful question quality. However, larger $\lambda$ over-emphasizes rubric scores relative to learnability, encouraging questions that are more complex but less matched to the current reasoner's capability. We therefore set $\lambda=0.2$ to balance quality guidance with learnability for proposer training.

\begin{table}[!htbp]
\centering
\caption{Performance of \oursolution{} across different base models.}
\label{tab:model_generalization}
\setlength{\tabcolsep}{2.0pt}
\renewcommand{\arraystretch}{1.18}
\resizebox{0.98\textwidth}{!}{%
\begin{tabular}{p{0.18\textwidth}ccccccc>{\columncolor{AvgHighlight}}c}
\toprule
\multirow{2}{*}{\textbf{Method}}
& \multicolumn{4}{c}{\textbf{Expert Medical Reasoning}}
& \multicolumn{1}{c}{\textbf{Clin. Case}}
& \multicolumn{2}{c}{\textbf{Science QA}}
& \cellcolor{white} \\
\cmidrule(lr){2-5} \cmidrule(lr){6-6} \cmidrule(lr){7-8}
& MedXpert
& MB-OP4
& MB-OP5
& MedQA
& LiveClin
& MMLU.
& GPQA
& Avg. \\
\midrule
\multicolumn{9}{c}{\textbf{ Qwen3-4B-Base}} \\
SFT (Warm-Start)
& 16.8 & 65.6 & 58.4 & 72.4 & 57.3 & 66.5 & 43.5 & 54.4 \\
R-Zero~\citep{r-zero}
& 16.3 & 65.5 & 59.4 & 73.8 & 59.4 & 68.4 & 44.2 & 55.3 \\
SPICE~\citep{spice}
& 17.6 & 66.4 & 60.9 & 77.6 & 61.7 & 70.7 & 45.5 & 57.2 \\
DARC~\citep{DARC}
& 18.4 & 67.1 & 60.8 & 78.4 & 62.6 & 71.4 & \textbf{47.8} & 58.1 \\
\textbf{\oursolution{}}
& \textbf{19.3} & \textbf{67.6} & \textbf{61.6} & \textbf{79.1}
& \textbf{64.4}
& \textbf{72.2} & 47.4
& \textbf{58.8} \\
\midrule
\multicolumn{9}{c}{\textbf{ OctoThinker-3B-Hybrid-Base}} \\
SFT (Warm-Start)
& 12.0 & 45.8 & 38.2 & 57.0 & 48.9 & 45.5 & 28.2 & 39.4 \\
R-Zero~\citep{r-zero}
& 10.5 & 44.5 & 39.4 & 55.4 & 49.6 & 45.1 & 28.4 & 39.0 \\
SPICE~\citep{spice}
& 12.9 & 47.2 & 41.7 & 58.5 & 50.8 & 47.1 & 29.8 & 41.1 \\
DARC~\citep{DARC}
& 12.6 & 47.9 & 40.1 & 58.0 & 51.3 & 46.2 & 28.9 & 40.7 \\
\textbf{\oursolution{}}
& \textbf{13.5} & \textbf{49.1} & \textbf{41.8} & \textbf{60.8}
& \textbf{51.8}
& \textbf{47.8} & \textbf{30.2}
& \textbf{42.1} \\
\bottomrule
\end{tabular}%
}
\end{table}

\subsection{Robustness Across Model Scales and Families}
Table~\ref{tab:model_generalization} examines whether the benefits of \oursolution{} extend beyond Qwen3-8B-Base. On Qwen3-4B-Base, \oursolution{} improves the average score from 54.4\% to 58.8\%, a gain of 4.4 points, and outperforms the strongest self-play baseline, DARC, by 0.7 points on average. On OctoThinker-3B-Hybrid-Base, \oursolution{} again achieves the best average performance, improving the warm-start model from 39.4\% to 42.1\%, a gain of 2.7 points. These results indicate that \oursolution{} is not tied to a specific base model and remains effective across different model scales and families.

\section{Conclusion}
In this paper, we show that directly applying document-grounded self-play RL to medical corpora is limited by two failure modes: \emph{knowledge-point fixation} and \emph{difficulty-quality mismatch}. To address these issues, we introduce \oursolution{}, which makes the clinical focus of question generation explicit through knowledge-point conditioning and augments proposer optimization with rubric-based quality feedback from the reasoner. Across seven medically related benchmarks, \oursolution{} consistently improves the warm-start model and outperforms existing self-play baselines, with especially strong gains on clinical case reasoning. Further ablations and generalization experiments support the robustness of these gains. These results demonstrate that \oursolution{} provides a scalable way to turn clinical case reports into continual training signals for improving medical reasoning.

\section*{Acknowledgments}
This paper is partially supported by the Hong Kong Research Grants Council grant R1018-25F.

{
\small
\bibliographystyle{unsrtnat}
\bibliography{main}
}
\newpage
\section*{\centering Appendix Contents}
\vspace{-1em}
\startcontents[sections]
\appendix
\printcontents[sections]{}{1}{}

\newpage

\section{Related Work}

\paragraph{RL for LLM Post-Training.}
RL has become increasingly important in LLM post-training, especially for improving reasoning in domains where rewards can be automatically verified~\citep{math_reasoning,kimi-k2,yao2026diversity,yao2026slat,wang2026learning}. Recent representative works, such as OpenAI-o1~\citep{openaio1} and DeepSeek-R1~\citep{dpskr1}, rely on large amounts of synthesized or curated training data with verifiable rewards, enabling effective RL training but often requiring substantial effort for data construction, verification, and filtering. Another line of work explores more weakly supervised or unsupervised alternatives, deriving training signals either from the model's own prior, such as majority voting~\citep{ttrl} and semantic similarity~\citep{empo}, or from unlabeled text~\citep{RLP,golden_goose,qstar}. The former can be brittle when model confidence is misaligned with correctness~\citep{unsupervised_rl}, while the latter seeks to extract supervision directly from raw text and thereby provides an alternative source of training signal. For example, Golden Goose~\citep{golden_goose} extends the fill-in-the-middle paradigm to construct supervision from document context, while Quiet-STaR~\citep{qstar} and RLPT~\citep{RLPT} optimize future tokens as supervision targets. These developments motivate our setting: medicine contains abundant high-quality unlabeled clinical text, making it a natural domain for studying whether self-play RL can convert real clinical corpora into useful supervision for medical reasoning.

\paragraph{RL for Medical Reasoning Models.}
A growing body of work applies RL post-training to medical reasoning and reports promising gains on medical benchmarks~\citep{huatuo-1,ehr-r1,medreason}. HuatuoGPT-o1~\citep{huatuo-1} is among the earliest efforts in this direction, using exam-style multiple-choice questions such as MedQA-USMLE~\citep{medqa-usmle} and MedMCQA~\citep{medmcqa} as verifiable training data. Fleming-R1~\citep{fleming-r1} and MedReason~\citep{medreason} further synthesize QA-style training instances by extracting medical entities and relations to construct medical knowledge graphs. AlphaMed~\citep{alphamed} builds a carefully curated medical RL dataset and highlights the importance of informative questions and diverse difficulty levels for effective training. MedCaseReasoning~\citep{MedcaseReasoning} identifies clinical case reports as a valuable source of medical reasoning data, but converts them into a static supervised dataset through extensive offline processing. Despite their encouraging progress, these methods predominantly rely on fixed offline supervision, including exam-style QA pairs, curated benchmarks, and synthesized reasoning datasets. In contrast, \oursolution{} treats clinical case reports not as a source of static annotations, but as grounding documents in an adaptive self-play loop that dynamically generates training questions for the current reasoner.

\paragraph{Self-Play RL.}
Self-play has long been a central paradigm in RL, with early successes in games such as AlphaZero~\citep{alphazero}, where models improve by repeatedly competing against themselves and refining their policies over iterations. Recent work extends this idea to LLM post-training. Absolute Zero~\citep{absolutezero} and R-Zero~\citep{r-zero} are among the earliest attempts to cast post-training as a co-evolutionary process between a problem proposer drawing on internal knowledge and a solver, enabling continual improvement without manually curated datasets. Subsequent methods introduce additional components into this loop. SPICE~\citep{spice} grounds problem generation in unlabeled text, while WIST~\citep{WIST} explores tree-structured self-play for acquiring domain-relevant knowledge from the open web. Despite this progress, existing self-play methods leave two limitations that are especially important in document-grounded medical reasoning. First, they typically do not explicitly control coverage over distinct knowledge points within a source document, which can cause the proposer to repeatedly anchor on a small subset of salient clues. Second, proposer optimization is still driven primarily by solver performance, which reflects difficulty but provides limited supervision over whether a question is clinically well-formed and useful for RL training. Motivated by these limitations, \oursolution{} addresses document-grounded medical self-play through knowledge-conditioned question generation and rubric-based quality feedback.

\section{Details of Visualization}
\label{apdx:visual}

\paragraph{Document Coverage Ratio.}
To measure how broadly generated questions cover the source case report, we first extract a set of fine-grained clues from each document $d$, denoted as
$\mathcal{C}_d=\{c_1,\ldots,c_{|\mathcal{C}_d|}\}$. These clues are more fine-grained than the knowledge points used by \oursolution{} and correspond to self-contained clinical facts, findings, interventions, outcomes, or mechanistic statements. We extract these clues once for each document using Qwen3-32B~\citep{qwen3}, obtaining an average of 48.3 clues per document.

Given a generated question $q$ for document $d$, we use Qwen3-32B to identify the subset of clues in $\mathcal{C}_d$ that are explicitly used by the question, denoted as
$\mathcal{C}^{\mathrm{used}}_{q,d}\subseteq \mathcal{C}_d$. At proposer-update step $u$, the document coverage ratio is computed by aggregating the unique clues covered by all valid questions generated from the same document up to that step, and then averaging the resulting document-level coverage ratios:
\begin{equation}
\mathrm{DCR}(u)
=
\frac{1}{|\mathcal{D}|}
\sum_{d\in\mathcal{D}}
\frac{
\left|\bigcup_{q\in \mathcal{Q}_d^{\le u}} \mathcal{C}^{\mathrm{used}}_{q,d}\right|
}{
|\mathcal{C}_d|
},
\label{eq:dcr}
\end{equation}
where $\mathcal{Q}_d^{\le u}$ denotes the set of valid questions generated from document $d$ up to proposer-update step $u$. Invalid questions are excluded before computing the ratio, and repeated use of the same clue within a document is counted only once.

\paragraph{Question Distribution Visualization.}
To examine the diversity of questions generated during self-play, we visualize the semantic distribution of questions produced by SPICE and \oursolution{}. Since the full set of generated questions is large, we randomly sample 100 source documents and analyze all questions generated from these documents. For each question, we concatenate the question lead-in and answer options, and encode the resulting text with Qwen3-Embedding-8B~\citep{qwen3embedding} to obtain semantic embeddings. We then compute pairwise cosine distances over the pooled question set from both methods and apply classical multidimensional scaling (MDS)~\citep{mds} to project the questions into two dimensions. Note that questions from SPICE and \oursolution{} are embedded and projected jointly rather than separately, so their relative positions in the MDS plane reflect semantic dispersion and overlap under the same embedding model and distance metric. The resulting visualization is shown in Figure~\ref{fig:pitfall_overview}(b).

\paragraph{Quality Distribution of Intermediate-Difficulty Questions.}
To compare the quality of questions generated by SPICE and \oursolution{}, we collect all proposer-generated questions throughout self-play training. We then select intermediate-difficulty questions, defined as questions for which the reasoner achieves an empirical answer accuracy of 0.5. This yields 7,915 questions for SPICE and 9,152 questions for \oursolution{}. From each set, we randomly sample 5,000 questions for evaluation. We use DeepSeek-V3.2~\citep{dpsk-v32} to score these questions with the same rubric prompt used in our framework. The rubric contains three dimensions: clue integration, reasoning depth, and option quality. We compute the overall quality score by averaging the three rubric dimensions and linearly normalizing the result to $[0,1]$. The final quality-score distributions are shown in Figure~\ref{fig:pitfall_overview}(c).

\section{Implementation Details}
\label{apdx:detail}

\subsection{Overview}
\label{apdx:detail:overview}
This section provides implementation details for \oursolution{}. We begin by describing the construction of the unlabeled clinical case-report corpus from the internet, including document-level filtering and preprocessing (Section~\ref{apdx:data}). We then describe how we extract a fixed set of document-grounded knowledge points for each retained report (Section~\ref{apdx:detail:kg_extraction}). During self-play, the proposer conditions on both the full case report and a selected knowledge point to generate a multiple-choice question, a supporting rationale, and a target answer (Section~\ref{apdx:detail:question_generation}). The reasoner evaluates each valid question using answer-based learnability feedback and rubric-based quality feedback (Section~\ref{apdx:detail:rubrics}). Finally, we detail the alternating self-play training procedure for updating the proposer and the reasoner (Section~\ref{apdx:detail:training_procedure}).

\definecolor{KPGreenBg}{HTML}{E7F1E4}
\definecolor{KPGreenBd}{HTML}{AFC6A8}
\definecolor{KPYellowBg}{HTML}{F6F0DE}
\definecolor{KPYellowBd}{HTML}{D1C092}
\definecolor{KPBlueBg}{HTML}{E8EAF8}
\definecolor{KPBlueBd}{HTML}{ADB3D8}
\definecolor{KPRedBg}{HTML}{F7E7E5}
\definecolor{KPRedBd}{HTML}{D3AAA5}

\newcommand{\omitted}{\textit{\textcolor{gray}{\texttt{<omitted parts>}}}}
\newcommand{\kptitle}[1]{\textbf{\strut #1}\par\vspace{2pt}}

\newcommand{\hlgreen}[1]{\begingroup\setlength{\fboxsep}{1pt}\colorbox{KPGreenBg}{#1}\endgroup}
\newcommand{\hlyellow}[1]{\begingroup\setlength{\fboxsep}{1pt}\colorbox{KPYellowBg}{#1}\endgroup}
\newcommand{\hlblue}[1]{\begingroup\setlength{\fboxsep}{1pt}\colorbox{KPBlueBg}{#1}\endgroup}
\newcommand{\hlred}[1]{\begingroup\setlength{\fboxsep}{1pt}\colorbox{KPRedBg}{#1}\endgroup}

\tcbset{
  kpbox/.style={
    boxrule=0.45pt,
    arc=0.9mm,
    left=5pt,right=5pt,top=4pt,bottom=4pt,
    valign=top,
    equal height group=caseplaykp
  }
}

\begin{figure*}[!htbp]
\centering
\begin{tcolorbox}[
  enhanced,
  width=\textwidth,
  colback=white,
  colframe=MedBlueLine,
  boxrule=0.7pt,
  arc=1.2mm,
  left=8pt,right=8pt,top=8pt,bottom=8pt
]
\footnotesize
\raggedright

\begin{center}
{\large\bfseries Clinical Case Report}
\end{center}

\vspace{4pt}
A 35-year-old White woman with hypertension and primary hyperaldosteronism presented with intermittent abdominal pain and chronic diarrhea for the past 6 years. She had a few episodes of small rectal bleeding. Review of systems was notable for heartburn and severe pain in her entire spine. The patient denied vomiting, weight loss, or fevers. Family history was notable for ulcerative colitis in her mother. \omitted

\vspace{4pt}
On physical examination, the patient had temperature 37.2°C, pulse rate 75, and blood pressure 140/100 with oxygen saturation 100\% on room air. There was mild epigastric tenderness on deep palpation. Initial laboratory examination showed a leukocyte count of 11.1 $\times$ 10$^{3}$/\textmu L (70\% neutrophils,  \hlgreen{1\% eosinophils, absolute eosinophil count 0.07 $\times$} \hlgreen{ 10$^{3}$/\textmu L} ), hemoglobin 13.2 $\times$ 10$^{3}$/\textmu L, platelet count 350 $\times$ 10$^{3}$/\textmu L, C-reactive protein 10 mg/dL, blood urea nitrogen 14 mg/dL, serum creatinine 0.76 mg/dL, and vitamin D 25 mg/dL.

\vspace{4pt}
An esophagogastroduodenoscopy and colonoscopy at an outside clinic 8 months ago
\hlyellow{appeared normal}.
Pathology was notable for
\hlyellow{intramucosal eosinophil at 53 eos/hpf on duodenal biopsy and 68 eos/hpf on gastric antrum biopsy}.
Helicobacter pylori was absent. On terminal ileum biopsy, there were
\hlgreen{120 intramucosal eos/hpf}.
Other random colon biopsies showed
\hlgreen{increased eosinophils of 87 eos/hpf}
and occasional noncaseating granulomas. However, there were no crypt abscesses or inflammatory cells in the lamina propria. \omitted~

\vspace{4pt}
The patient was started on
\hlblue{budesonide 6 mg daily with mild improvement},
but her symptoms relapsed with taper to
\hlblue{budesonide 3 mg daily}.
Oral prednisone was contraindicated due to primary hyperaldosteronism. Given
\hlblue{steroid dependence}
and concomitant arthritis, she underwent induction with adalimumab and was bridged to adalimumab 40 mg every 14 days. \omitted~

\vspace{4pt}
Eventually, the patient had
\hlred{complete resolution of her GI and joint symptoms on an adalimumab 40 mg weekly dose}
after budesonide was tapered off. Five months after adalimumab initiation, her inflammatory markers were normalized \omitted~The biopsies showed nonspecific antrum erosion and nonspecific mild prominence of eosinophils in the cecum; \hlred{the rest of the biopsies were normal.}

\vspace{8pt}
\hrule
\vspace{8pt}

\begin{center}
\begin{tabular}{p{0.47\textwidth} p{0.47\textwidth}}
\begin{tcolorbox}[kpbox, colback=KPGreenBg, colframe=KPGreenBd]
\raggedright
\kptitle{Knowledge Point 1}
In eosinophilic gastroenteritis and colitis, peripheral absolute eosinophil count may remain within normal limits despite dense tissue eosinophilic infiltration.
\end{tcolorbox}
&
\begin{tcolorbox}[kpbox, colback=KPYellowBg, colframe=KPYellowBd]
\raggedright
\kptitle{Knowledge Point 2}
Eosinophilic gastroenteritis and colitis can be diagnosed based on increased intramucosal eosinophils on biopsy even when endoscopic visualization appears normal.
\end{tcolorbox}
\\[6pt]
\begin{tcolorbox}[kpbox, colback=KPBlueBg, colframe=KPBlueBd]
\raggedright
\kptitle{Knowledge Point 3}
Budesonide treatment may lead to initial improvement in eosinophilic gastroenteritis and colitis, but symptoms can relapse upon dose reduction, indicating steroid dependence.
\end{tcolorbox}
&
\begin{tcolorbox}[kpbox, colback=KPRedBg, colframe=KPRedBd]
\raggedright
\kptitle{Knowledge Point 4}
Adalimumab can induce clinical and histologic remission in adult patients with steroid-refractory eosinophilic gastroenteritis and colitis.
\end{tcolorbox}
\end{tabular}
\end{center}

\end{tcolorbox}
\caption{Case study of knowledge-point extraction from a clinical case report. }
\label{fig:caseplay_case_study}
\end{figure*}

\subsection{Data Preparation}
\label{apdx:data}
We collect open-access clinical case reports from the internet published between 2020 and 2025. We apply a three-stage filtering pipeline to ensure that each retained document contains sufficient text-grounded clinical evidence for question generation. \emph{First}, we parse the raw XML files and discard reports that cannot be parsed, lack full text, have incomplete formatting, or are too short (fewer than 500 tokens). \emph{Second}, we perform corpus-level filtering and deduplication. We first remove duplicate entries by document ID, then remove near-duplicate reports using n-gram similarity. We also use regular expressions to count figures and tables, excluding reports dominated by visual content because our experiments focus on text-grounded question generation. To reduce benchmark leakage, we apply both document-level and fine-grained benchmark deduplication. At the document level, we remove any training report whose document ID appears in LiveClin. At the fine-grained level, we split each report into chunks and compare each chunk against every benchmark question using n-gram similarity. If any chunk is detected as a near duplicate of a benchmark question, we discard the entire report from the training corpus. \emph{Third}, for the remaining reports, we remove publication-related and medically irrelevant sections, including acknowledgements, funding statements, author contributions, conflicts of interest, references, and other metadata. When section information is available, we prioritize clinically relevant sections such as case presentation, diagnosis, treatment, outcome, and discussion. Finally, we retain document content that fits within the maximum input length of 4,096 tokens. After preprocessing, the final corpus contains $|\mathcal{D}| = 4{,}281$ case reports. These documents are used only as unlabeled source documents during self-play.

Before self-play, we construct a small warm-start dataset to stabilize instruction following, role-specific behavior, and structured output formatting. Specifically, we randomly sample 20\% of the retained training reports and use DeepSeek-V3.2~\citep{dpsk-v32} to generate one proposer question for each extracted knowledge point. For each generated question, we further sample two correct reasoner responses. This process yields 3,384 proposer examples and 6,768 reasoner examples, which are used for SFT warm-up before self-play training.

\subsection{Knowledge-Point Extraction}
\label{apdx:detail:kg_extraction}
Before self-play training, we extract a set of document-grounded knowledge points $K(d)=\{k_1,\dots,k_M\}$ for each case report $d\in\mathcal{D}$. Each knowledge point is intended to capture a clinically meaningful and case-specific unit that is explicitly supported by the document, such as a distinctive clinical finding, a treatment-response pattern, or a report-supported disease association or mechanism. The extraction procedure is guided by the prompt in Table~\ref{tab:kg_extraction_prompt} and follows three principles: (a) every knowledge point must be grounded in evidence from the case report, without introducing unsupported facts from external medical knowledge; (b) knowledge points extracted from the same report should be mutually non-overlapping and cover different clinical aspects; and (c) we keep at most $M=4$ knowledge points for each case report in our main experiments. The resulting knowledge points are fixed throughout RL training, which avoids introducing additional non-stationarity into \oursolution{}. Figure~\ref{fig:caseplay_case_study} presents an example case report together with its extracted knowledge points. As shown in the figure, non-overlapping knowledge points align with different parts of the original report, helping the proposer cover a broader range of source content during question generation.


\newcommand{\accpill}[2]{%
  {\scriptsize\bfseries Average Accuracy: #2}%
}

\newcommand{\qhead}[3]{%
  \noindent
  {\bfseries #1}
  \hfill
  \par\vspace{2pt}
}

\newcommand{\qsection}[1]{%
  \vspace{3pt}
  {\bfseries #1}\par\vspace{1pt}
}

\newcommand{\optline}[2]{%
  \noindent\hangindent=1.35em\hangafter=1
  \textbf{#1)} #2\par
}

\newcommand{\optsfour}[4]{%
  \vspace{1pt}
  \optline{A}{#1}
  \optline{B}{#2}
  \optline{C}{#3}
  \optline{D}{#4}
}

\newcommand{\optsfive}[5]{%
  \vspace{1pt}
  \optline{A}{#1}
  \optline{B}{#2}
  \optline{C}{#3}
  \optline{D}{#4}
  \optline{E}{#5}
}

\tcbset{
  qgridboxtop/.style={
    enhanced,
    boxrule=0.45pt,
    arc=0.9mm,
    left=5pt,right=5pt,top=4pt,bottom=4pt,
    height=0.218\textheight,
    valign=top,
    before upper={\raggedright\scriptsize\setlength{\parskip}{0pt}}
  },
  qgridboxbottom/.style={
    enhanced,
    boxrule=0.45pt,
    arc=0.9mm,
    left=5pt,right=5pt,top=4pt,bottom=4pt,
    height=0.235\textheight,
    valign=top,
    before upper={\raggedright\scriptsize\setlength{\parskip}{0pt}}
  }
}

\begin{figure*}[!t]
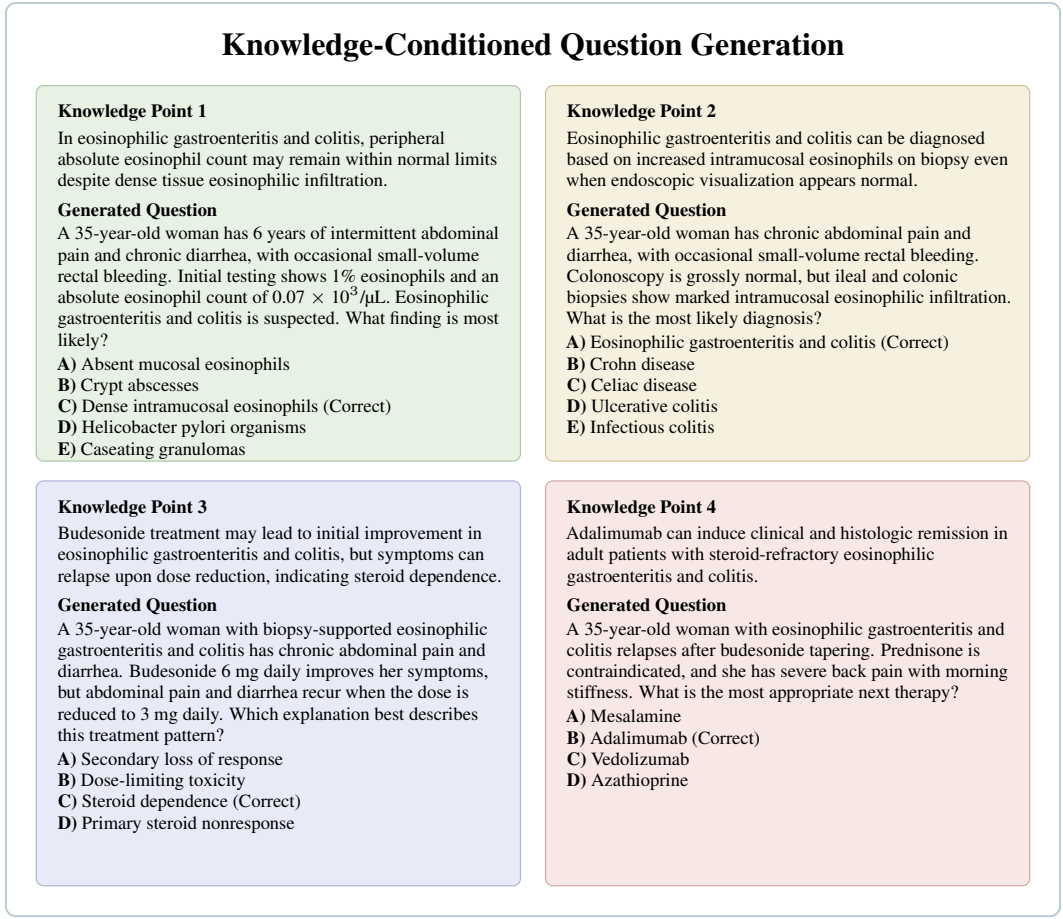

\centering
\begin{tcolorbox}[
  enhanced,
  width=\textwidth,
  colback=white,
  colframe=MedBlueLine,
  boxrule=0.7pt,
  arc=1.2mm,
  left=8pt,right=8pt,top=8pt,bottom=8pt
]

\begin{center}
{\large\bfseries Knowledge-Conditioned Question Generation}
\end{center}

\vspace{3pt}

\noindent
\begin{minipage}[t]{0.488\textwidth}
\begin{tcolorbox}[qgridboxtop, colback=KPGreenBg, colframe=KPGreenBd]
\qhead{Knowledge Point 1}{KPGreenBd}{}

In eosinophilic gastroenteritis and colitis, peripheral absolute eosinophil count may remain within normal limits despite dense tissue eosinophilic infiltration.

\qsection{Generated Question}
A 35-year-old woman has 6 years of intermittent abdominal pain and chronic diarrhea, with occasional small-volume rectal bleeding. Initial testing shows 1\% eosinophils and an absolute eosinophil count of 0.07 $\times$ 10$^3$/\textmu L. Eosinophilic gastroenteritis and colitis is suspected. What finding is most likely?

\optsfive
{Absent mucosal eosinophils}
{Crypt abscesses}
{Dense intramucosal eosinophils (Correct)}
{Helicobacter pylori organisms}
{Caseating granulomas}

\end{tcolorbox}
\end{minipage}
\hfill
\begin{minipage}[t]{0.488\textwidth}
\begin{tcolorbox}[qgridboxtop, colback=KPYellowBg, colframe=KPYellowBd]
\qhead{Knowledge Point 2}{KPYellowBd}{}

Eosinophilic gastroenteritis and colitis can be diagnosed based on increased intramucosal eosinophils on biopsy even when endoscopic visualization appears normal.

\qsection{Generated Question}
A 35-year-old woman has chronic abdominal pain and diarrhea, with occasional small-volume rectal bleeding. Colonoscopy is grossly normal, but ileal and colonic biopsies show marked intramucosal eosinophilic infiltration. What is the most likely diagnosis?

\optsfive
{Eosinophilic gastroenteritis and colitis (Correct)}
{Crohn disease}
{Celiac disease}
{Ulcerative colitis}
{Infectious colitis}

\end{tcolorbox}
\end{minipage}

\vspace{6pt}

\noindent
\begin{minipage}[t]{0.488\textwidth}
\begin{tcolorbox}[qgridboxbottom, colback=KPBlueBg, colframe=KPBlueBd]
\qhead{Knowledge Point 3}{KPBlueBd}{}

Budesonide treatment may lead to initial improvement in eosinophilic gastroenteritis and colitis, but symptoms can relapse upon dose reduction, indicating steroid dependence.

\qsection{Generated Question}
A 35-year-old woman with biopsy-supported eosinophilic gastroenteritis and colitis has chronic abdominal pain and diarrhea. Budesonide 6 mg daily improves her symptoms, but abdominal pain and diarrhea recur when the dose is reduced to 3 mg daily. Which explanation best describes this treatment pattern?

\optsfour
{Secondary loss of response}
{Dose-limiting toxicity}
{Steroid dependence (Correct)}
{Primary steroid nonresponse}

\end{tcolorbox}
\end{minipage}
\hfill
\begin{minipage}[t]{0.488\textwidth}
\begin{tcolorbox}[qgridboxbottom, colback=KPRedBg, colframe=KPRedBd]
\qhead{Knowledge Point 4}{KPRedBd}{}

Adalimumab can induce clinical and histologic remission in adult patients with steroid-refractory eosinophilic gastroenteritis and colitis.

\qsection{Generated Question}
A 35-year-old woman with eosinophilic gastroenteritis and colitis relapses after budesonide tapering. Prednisone is contraindicated, and she has severe back pain with morning stiffness. What is the most appropriate next therapy?

\optsfour
{Mesalamine}
{Adalimumab (Correct)}
{Vedolizumab}
{Azathioprine}

\end{tcolorbox}
\end{minipage}

\end{tcolorbox}

\caption{Examples of knowledge-conditioned question generation from the same clinical case report.}
\vspace{-5mm}
\label{fig:kp_question_acc_examples}
\end{figure*}

\subsection{Knowledge-Conditioned Question Generation}
\label{apdx:detail:question_generation}
As shown in Table~\ref{tab:proposer_prompt}, we concatenate the case report with the selected target knowledge point and ask the proposer to generate one self-contained USMLE-style multiple-choice question centered on that target. The prompt emphasizes four requirements: (a) the question must remain grounded in the source document; (b) it should focus on the selected knowledge point rather than test it through direct restatement; (c) it should require nontrivial medical reasoning, preferably integrating at least three distinct clues and involving a two-hop inference chain when possible; and (d) the answer options should be well-formed, with 4 to 8 clinically plausible options of the same semantic type and granularity.

We further require the proposer to return a structured output with five fields: \texttt{question\_type}, \texttt{exam\_question}, \texttt{options}, \texttt{correct\_answer}, and \texttt{knowledge\_and\_reasoning\_steps}. The \texttt{question\_type} field is chosen from \texttt{diagnosis}, \texttt{mechanism}, \texttt{finding}, \texttt{explanation}, and \texttt{treatment}; \texttt{exam\_question} contains the self-contained vignette and lead-in; \texttt{options} contains the answer candidates; \texttt{correct\_answer} gives the label of the gold option; and \texttt{knowledge\_and\_reasoning\_steps} summarizes the intended clue-to-inference chain. Outputs with missing fields, malformed options, non-contiguous option labels, ambiguous correct answers, or unparseable structure are marked invalid. Invalid outputs receive a fixed format penalty $r^{\mathrm{format}}$ during proposer optimization and are excluded from reasoner training. Figure~\ref{fig:kp_question_acc_examples} presents examples of questions generated from the same source document under different knowledge-point conditions.

\subsection{Rubric-Based Judge with the Reasoner}
\label{apdx:detail:rubrics}

In \oursolution{}, we use a rubric-based judge prompt to evaluate the quality of each proposer-generated question, providing richer feedback signals for proposer optimization. The full prompt is shown in Table~\ref{tab:rubric_prompt}. The judge is given the generated question, the target answer, and the proposer-provided reasoning chain, and is asked to assess the question along three dimensions:
\begin{itemize}[leftmargin=1.5em,itemsep=0.2em,topsep=0.2em]
    \item \textbf{Clue integration} measures whether answering the question requires synthesizing multiple non-redundant, case-derived clues, rather than relying on a single giveaway fact or a simple knowledge-matching shortcut. It discourages the proposer from generating overly direct questions that can be solved through surface-level clue recognition alone.
    \item \textbf{Reasoning depth} measures whether the question requires a necessary multi-step clinical inference process, rather than shallow fact retrieval or post-hoc explanation. Importantly, the rubric distinguishes between using multiple clues and requiring multiple reasoning hops: several clues pointing directly to the same answer do not by themselves imply deep reasoning.
    \item \textbf{Option quality} measures whether the option set is homogeneous, clinically plausible, and sufficiently discriminative, such that the correct answer is most reliably identified through case-based reasoning rather than wording cues or obviously weak distractors.
\end{itemize}

Each dimension is scored independently on a 1--5 scale according to the detailed criteria in Table~\ref{tab:rubric_prompt}. We then linearly normalize the resulting scores to $[0,1]$ and use them as separate quality reward channels in proposer optimization.

\begin{algorithm}[t]
\caption{Training \oursolution{}}
\label{alg:caseplay}
\begin{algorithmic}[1]
\Require Proposer $\pi_\theta^P$, reasoner $\pi_\phi^R$, document corpus $\mathcal{D}$
\State Extract document-grounded knowledge points $\{K(d)\}_{d \in \mathcal{D}}$
\For{iteration $t = 1, \dots, T$}
    \State Initialize the question buffer $\mathcal{B}_t \leftarrow \emptyset$

    \State \textbf{// Phase I: Proposer update}
    \For{proposer update step $u = 1, \dots, U_P$}
        \State Sample a minibatch of documents $\mathcal{D}_u \subset \mathcal{D}$
        \For{each document $d \in \mathcal{D}_u$}
            \State Select rollout knowledge points $\{k_{d,j}\}_{j=1}^{N_p}$ according to $\pi^{\mathrm{sel}}(\cdot \mid K(d),j)$
            \State Sample candidate outputs $\{o_{d,j}=(q_{d,j},c_{d,j},y_{d,j})\}_{j=1}^{N_p}$ from $\pi_\theta^P(\cdot \mid d,k_{d,j})$
            \State Assign the format penalty $r^{\mathrm{format}}$ to invalid outputs and skip their evaluation
            \State Sample $N_{\mathrm{acc}}$ answers from $\pi_\phi^R(\cdot \mid q_{d,j})$ without access to $d$
            \State Estimate answer accuracy against $y_{d,j}$ and compute $r_{d,j}^{\mathrm{learn}}$ by Eq.~(\ref{eq:learnability_reward})
            \State Evaluate generated questions with the rubric judge and obtain $(r_{d,j}^{\mathrm{clue}}, r_{d,j}^{\mathrm{depth}}, r_{d,j}^{\mathrm{option}})$
            \State Add generated question-answer pairs $\{(q_{d,j},y_{d,j})\}$ to question buffer $\mathcal{B}_t$
            \State Compute normalized proposer advantages $\{\hat{A}_{d,j}^P\}_{j=1}^{N_p}$ by Eq.~(\ref{eq:component_advantage})

        \EndFor
        \State Update $\pi_\theta^P$ on the current minibatch by Eq.~(\ref{eq:grpo_obj})
    \EndFor

    \State \textbf{// Phase II: Reasoner update}
    \For{reasoner update step $u = 1, \dots, U_R$}
        \State Sample a minibatch of valid question-answer pairs $\{(q_i,y_i)\}$ from $\mathcal{B}_t$
        \For{each sampled pair $(q_i,y_i)$}
            \State Sample $N_r$ responses $\{a_{i,j}\}_{j=1}^{N_r} \sim \pi_\phi^R(\cdot \mid q_i)$
            \State Assign rewards $\{r_{i,j}^R\}_{j=1}^{N_r}$ according to whether the final answer in $a_{i,j}$ matches $y_i$
            \State Compute group-normalized reasoner advantage $\{\hat{A}_{i,j}^R\}_{j=1}^{N_r}$
        \EndFor
        \State Update $\pi_\phi^R$ on the current minibatch by Eq.~(\ref{eq:grpo_obj})
    \EndFor
\EndFor
\end{algorithmic}
\end{algorithm}

\subsection{Self-Play Training}
\label{apdx:detail:training_procedure}
Algorithm~\ref{alg:caseplay} summarizes the full training procedure of \oursolution{}. We first extract a fixed set of document-grounded knowledge points for the training corpus (line 1). In each self-play iteration, we initialize an empty question buffer $\mathcal{B}_t$ (line 3) and then alternate between a proposer-update phase and a reasoner-update phase.

During the proposer-update phase (lines 4--17), we sample a minibatch of documents and, for each document, select rollout-specific knowledge points according to the knowledge-point selection strategy (lines 7--9). The proposer then generates $N_p$ candidate outputs conditioned on the document and the selected knowledge points. Invalid outputs receive the format penalty and are excluded from reasoner evaluation (line 10). For the remaining generated questions, the frozen reasoner samples answers without access to the source document and computes the learnability reward (lines 11--12). The same generated questions are also evaluated by the rubric judge to obtain quality rewards (line 13). We add the generated question-answer pairs to the question buffer and compute normalized proposer advantages from the reward components before updating the proposer with GRPO (lines 15--17).

During the reasoner-update phase (lines 19--28), we sample valid question-answer pairs from the question buffer (line 21). For each sampled question, the reasoner generates $N_r$ responses without access to the source document (line 23). Each response receives a binary reward according to whether its final answer matches the target answer, and the resulting rewards are group-normalized to obtain reasoner advantages (lines 24--25). The reasoner is also updated using GRPO (line 27).

\FloatBarrier
\clearpage

\begin{promptboxfloat}{%
\phantomsection
\refstepcounter{table}\label{tab:kg_extraction_prompt}%
Table~\thetable: Prompt for Knowledge Point Extraction}
You are a medical knowledge-point extraction model working on a single case report. Your task is to extract multiple high-value knowledge points directly from the case report. You must therefore identify clinically useful knowledge points from scratch, using only the case report content.

(*@{\bfseries\sffamily \# Input}@*)

(*@{\bfseries\sffamily \#\# Document}@*)
(*@{\bfseries\sffamily\textcolor{MedBlue}{<Case Report>}}@*)

(*@{\bfseries\sffamily \# Goal}@*)
Before writing the final output, first read the report and identify clinically meaningful findings, tests, management details, complications, and outcomes that are explicitly supported. Then propose candidate knowledge points, remove overlapping candidates, and keep the 4 strongest remaining candidates based on case-specificity and evidence strength that:
1. are directly supported by the report text
2. are clinically meaningful
3. are mutually non-overlapping
4. reflect distinct reasoning targets, findings, or management implications from the same case
Never invent unsupported facts just to increase the number of points.

(*@{\bfseries\sffamily \# Hard Grounding Rules}@*)
You must obey all of the following:
1. Use the case report as the primary source of truth.
2. Every generated point must be supported by explicit content in the report text.
3. Do not use outside medical knowledge to add facts that are not stated in the report.
4. Do not turn weak hints into confident claims.
5. Do not produce generic textbook facts that could fit many unrelated cases.
6. Do not produce image-only points unless the imaging finding is explicitly described in text.
7. Do not produce publication trivia, ethics details, or administrative details.
8. Do not use "this is the first reported case" style novelty as a knowledge point unless it creates a real clinical teaching point, which is usually not the case.

(*@{\bfseries\sffamily \# Non-Overlap Rules}@*)
The generated points must be distinct from each other. Treat the following as duplicates and avoid:
- the same diagnosis phrased in two different ways
- the same clinical clue boundary rewritten in another form
- a broad point and a narrower restatement of the same point

(*@{\bfseries\sffamily \# Preferred Knowledge Point Types}@*)
- a clinically important diagnosis stated or strongly established in the report
- a distinctive clinical, laboratory, imaging, pathology, or procedural finding
- a management or treatment response detail that is specific and clinically useful
- a complication, progression pattern, or follow-up finding explicitly shown in the report
- a disease association or mechanism only if the report itself supports it

(*@{\bfseries\sffamily \# Writing Rules}@*)
- Write all output in English.
- Each knowledge point should be one concise sentence or clause.
- Keep it specific, case-grounded, and teaching-oriented.
- Avoid repeating the same disease across multiple points unless the clinical boundary is clearly different.
- Prefer wording that would still be useful as retrieval-oriented supervision later.

(*@{\bfseries\sffamily \# Output Format}@*)
Return exactly one JSON object with this schema:
{
  "generated_knowledge_points": [
    "knowledge point 1",
    "knowledge point 2",
    "..."
  ]
}


\end{promptboxfloat}

\begin{promptboxfloat}{%
\phantomsection
\refstepcounter{table}\label{tab:proposer_prompt}%
Table~\thetable: Prompt for Proposer Question Generation}
Your task is to write one self-contained USMLE-style multiple-choice question from the case document for student learning. You will be given a case document and one target knowledge point. Your job is to use the provided knowledge point to guide the question you write, so that solving the question helps reinforce that target. When deciding what to ask:
- keep the focus on the provided target knowledge point
- adjust the question angle and distractors so the item clearly teaches and reinforces that target
- aim to teach, reinforce, and clarify the target knowledge point rather than only test it mechanically

(*@{\bfseries\sffamily \# Input}@*)

(*@{\bfseries\sffamily \#\# Document}@*)
(*@{\bfseries\sffamily\textcolor{MedBlue}{<Case Report>}}@*)

(*@{\bfseries\sffamily \#\# Target Knowledge Point}@*)
(*@{\bfseries\sffamily\textcolor{MedBlue}{<Target Knowledge Point>}}@*)

(*@{\bfseries\sffamily \# Requirements}@*)
(*@{\bfseries\sffamily \#\# Grounding}@*)
- The correct answer must be supported by the document.
- Do not invent decisive diagnoses, mechanisms, findings, timelines, management steps, or outcomes.
- The question must be solvable without mentioning the source article.

(*@{\bfseries\sffamily \#\# Reasoning Quality}@*)
- Prefer items that require integrating at least (*@{\bfseries\rmfamily 3 distinct clues}@*).
- Prefer a real (*@{\bfseries\rmfamily 2-hop or 3-hop}@*) reasoning chain when possible.
- Avoid direct restatement, buzzword-only items, trivia, and image-related questions.

(*@{\bfseries\sffamily \#\# Question Type}@*)
Choose exactly one `question_type` from `diagnosis`, `mechanism`, `finding`, `explanation`, and `treatment`.

(*@{\bfseries\sffamily \#\# Options}@*)
- Use (*@{\bfseries\rmfamily 4 to 8}@*) options when possible.
- Use contiguous labels starting from `A`.
- All options must be the same semantic type and level.
- Include plausible near-neighbor distractors.
- Do not use duplicate option text.

(*@{\bfseries\sffamily \# Output Format}@*)
1. First write the reasoning content inside a <think>...</think> block.
2. Immediately after </think>, output a JSON object.
3. Do not output markdown, code fences, or any extra text outside the think block and the JSON object.
4. Strictly follow the JSON schema below:
{
  "question_type": "diagnosis | mechanism | finding | explanation | treatment",
  "exam_question": "self-contained vignette and lead-in only",
  "options": {
    "A": "option text",
    "B": "option text",
    "..."
  },
  "correct_answer": "label of the correct answer",
  "knowledge_and_reasoning_steps": [
    "clue -> inference",
    "..."
  ]
}

\end{promptboxfloat}

\begin{promptboxfloat}{%
\phantomsection
\refstepcounter{table}\label{tab:rubric_prompt}%
Table~\thetable: Prompt for Rubric Judge}
You are an expert USMLE item quality reviewer with deep knowledge of clinical reasoning assessment and NBME item-writing standards.

(*@{\bfseries\sffamily \# Evaluation Dimensions}@*)
(*@{\bfseries\sffamily \#\# 1. Clue Integration}@*)
- (*@{\bfseries\rmfamily Definition:}@*) Solving the item requires integrating multiple clinically meaningful clues rather than matching a single giveaway fact.
- (*@{\bfseries\rmfamily Scoring Criteria:}@*)
  - (*@{\bfseries\rmfamily 1:}@*) Almost entirely single-clue; extra details are decorative and do not materially affect choices.
  - (*@{\bfseries\rmfamily 2:}@*) More than one clue is present, but one dominant clue still does nearly all of the work and the others merely confirm it.
  - (*@{\bfseries\rmfamily 3:}@*) At least two meaningful clues must be combined, but the synthesis is still fairly shallow or straightforward.
  - (*@{\bfseries\rmfamily 4:}@*) Multiple distinct clues must be integrated to narrow the differential or choose among close alternatives; no single clue is sufficient.
  - (*@{\bfseries\rmfamily 5:}@*) The answer depends on rich, case-specific synthesis across several clues, including supporting and discriminating details, and would become unstable if any major clue were removed.

(*@{\bfseries\sffamily \#\# 2. Reasoning Depth}@*)
- (*@{\bfseries\rmfamily Definition:}@*) Judge whether the correct answer requires a multi-hop reasoning chain, not merely multiple clues pointing directly to one answer.
- (*@{\bfseries\rmfamily Scoring Criteria:}@*)
  - (*@{\bfseries\rmfamily 1:}@*) One-hop reasoning: long vignette, but the answer is still selected from one dominant clue.
  - (*@{\bfseries\rmfamily 2:}@*) Appears multi-step on the surface, but one step is weak, optional, or post-hoc explanation rather than necessary reasoning.
  - (*@{\bfseries\rmfamily 3:}@*) Clear and necessary 2-hop reasoning: clues must first support an intermediate clinical judgment, and that judgment is then needed to choose the answer.
  - (*@{\bfseries\rmfamily 4:}@*) Strong 2-hop or genuine 3-hop reasoning: multiple clues support a necessary intermediate conclusion, which then supports option-level discrimination against strong distractors.
  - (*@{\bfseries\rmfamily 5:}@*) Clear, necessary, and non-compressible 3-hop reasoning: each hop is required, each hop depends on the prior one, and at least one strong distractor survives early reasoning but is eliminated only by the full chain.

(*@{\bfseries\sffamily \#\# 3. Option Quality}@*)
- (*@{\bfseries\rmfamily Definition:}@*) The option set supports fair one-best-answer discrimination through comparable, plausible, and clinically coherent alternatives. The item may contain  4 to 8 options labeled from `A` onward. Penalize this dimension only when the added options reduce homogeneity, plausibility, or one-best-answer fairness.
- (*@{\bfseries\rmfamily Scoring Criteria:}@*)
  - (*@{\bfseries\rmfamily 1:}@*) Most distractors are weak, implausible, or cueing; one-best-answer discrimination is largely artificial.
  - (*@{\bfseries\rmfamily 2:}@*) Some distractors are plausible, but the set still has noticeable heterogeneity, stylistic cues, or obvious throwaway options.
  - (*@{\bfseries\rmfamily 3:}@*) Options are mostly on the same semantic level and at least one distractor is reasonably strong, but the correct answer still stands out too easily.
  - (*@{\bfseries\rmfamily 4:}@*) Options are homogeneous and clinically plausible, with nearby distractors that require real case-based reasoning to eliminate.
  - (*@{\bfseries\rmfamily 5:}@*) The option set is exceptionally tight: multiple distractors remain genuinely attractive until the full case is reasoned through, and the correct answer wins without wording cues or weak-option shortcuts.

(*@{\bfseries\sffamily \# Counting Criteria}@*)
- (*@{\bfseries\rmfamily Clue:}@*) A clue is an independent, non-redundant piece of case information that changes answer discrimination. If two details provide the same clinical signal, count them as one clue, not two.
- (*@{\bfseries\rmfamily Hop:}@*) A hop is a necessary intermediate inference step that produces a new clinical conclusion needed for the next step or the final answer. Do not count post-hoc explanation or optional intermediate wording as an extra hop.

(*@{\bfseries\sffamily \# Evaluation Rules}@*)
- Score each dimension separately. Do not let strength in one compensate for weakness in another.
- Do not confuse "many clues" with "many hops." Multiple clues that all point directly to one diagnosis do not by themselves justify a high `reasoning_depth` score.
- Treat 4 to 8 options as acceptable; judge the quality of the set, not the count alone.

(*@{\bfseries\sffamily \# Output Format}@*)
Return exactly one JSON object with this schema:
{
  "scores": {
    "clue_integration": { "score": 1, "reason": "Evidence-based rationale" },
    "reasoning_depth": { "score": 1, "reason": "Evidence-based rationale" },
    "option_set_quality": { "score": 1, "reason": "Evidence-based rationale" }
  }
}
(*@{\bfseries\sffamily \# Input}@*)
(*@{\bfseries\sffamily \#\# Questions}@*)
(*@{\bfseries\sffamily\textcolor{MedBlue}{<Generated Question>}}@*)
(*@{\bfseries\sffamily \#\# Answers}@*)
(*@{\bfseries\sffamily\textcolor{MedBlue}{<Generated Target Answer>}}@*)
(*@{\bfseries\sffamily \#\# Reasoning chain}@*)
(*@{\bfseries\sffamily\textcolor{MedBlue}{<Generated Reasoning Chain>}}@*)

\end{promptboxfloat}

\FloatBarrier
\clearpage

\section{Experimental Setup}
\label{apdx:exp:detail_all}

\subsection{Comparison Baselines}
\label{apdx:exp:baselines}

We compare \oursolution{}~against two groups of baselines: \emph{existing medical reasoning models} and \emph{state-of-the-art self-play methods}. For medical reasoning models, we select three representative models, including:
\begin{itemize}
    \item \textbf{HuatuoGPT-o1~\citep{huatuo-1}} is among the first works to study how to improve medical reasoning through RL. It first uses SFT to elicit medical reasoning behavior and then applies RL with medical verifiable rewards to enhance reasoning performance.

    \item \textbf{AlphaMed~\citep{alphamed}} systematically studies how minimalist rule-based RL can improve model performance on medical reasoning benchmarks. It further highlights the importance of training data diversity for enhancing medical reasoning ability, making it a strong representative of medical RL post-training.

    \item \textbf{Fleming-R1~\citep{fleming-r1}} follows an offline data construction pipeline that first builds medical knowledge graphs and then synthesizes QA questions from the constructed graphs. The synthesized QA data are then used for RL training, leading to strong performance among medical reasoning models.
\end{itemize}

For self-play methods, we include one document-free self-play baseline and two recent document-grounded self-play baselines:
\begin{itemize}
    \item \textbf{R-Zero~\citep{r-zero}}:
    R-Zero is a zero-data self-play framework in which a proposer generates questions based on its own internal knowledge, and a reasoner learns from the generated questions. We include this baseline to examine whether self-play that relies only on the model's internal knowledge is sufficient for medical reasoning. In our implementation, we remove both document input and knowledge-point conditioning from the proposer, and also remove rubric-judge feedback. All other training details follow the original R-Zero setup as closely as possible under our backbone model and training budget.

    \item \textbf{SPICE~\citep{spice}}:
    SPICE is a document-grounded self-play method that uses unlabeled text to support challenger question generation. Unlike \oursolution{}, SPICE uses the same model as both proposer and reasoner, and updates them jointly within each batch. Following SPICE, we merge proposer-side and reasoner-side data for warm-start training, and then jointly sample and update the shared model during self-play. We keep the number of update examples matched to \oursolution{}~for a fair comparison.

    \item \textbf{DARC~\citep{DARC}}:
    DARC is a two-stage document-grounded self-play method that trains the challenger and reasoner separately. Following its original design, we first train the proposer with explicit difficulty targets using grounding and difficulty rewards, and then use the trained proposer to generate questions for reasoner training. The total amount of training data is matched to \oursolution{}~to ensure a fair comparison.
\end{itemize}

\subsection{Benchmarks}
\label{apdx:exp:benchmarks}

We evaluate \oursolution{}~on 7 medically related benchmarks covering expert-level medical reasoning, clinical case reasoning, and scientific QA. All benchmarks are evaluated in a multiple-choice format. During evaluation, we prompt the model to place its final answer in \texttt{\textbackslash boxed\{\}}, and use \texttt{math-verify}~\citep{math-verify} to extract the boxed answer and compare it with the ground-truth option. The details of each benchmark are as follows.

\begin{itemize}
    \item \textbf{MedXpert-QA}~\citep{medXpertQA} is an expert-level medical QA benchmark designed to evaluate advanced clinical knowledge and reasoning. In this paper, we evaluate on the MedXpert-Text subset, which contains 2,450 text-only multiple-choice questions.

    \item \textbf{MedBullets-OP4/OP5}~\citep{op5} consists of exam-style medical questions derived from MedBullets. We evaluate both the four-option and five-option variants, denoted as OP4 and OP5, respectively. Each variant contains 308 questions.

    \item \textbf{MedQA-USMLE}~\citep{medqa-usmle} is a widely used medical exam benchmark collected from United States Medical Licensing Examination-style questions, evaluating a model's ability to apply medical knowledge and clinical reasoning in realistic clinical scenarios. We use the standard test split, which contains 1,273 questions.
    
    \item \textbf{LiveClin}~\citep{liveclin} is a clinical case reasoning benchmark built from patient-centered clinical scenarios. It emphasizes realistic case-based reasoning, including symptom interpretation, diagnosis, and treatment decision-making. Since the original benchmark contains multimodal questions, we filter out questions that depend on figures or images, resulting in 1,450 text-only questions.

    \item \textbf{MMLU-Pro}~\citep{mmlu_pro} is a challenging extension of MMLU with harder questions and expanded answer choices. We evaluate the medically related Biology and Health subsets, which contain 717 and 818 questions, respectively, for a total of 1,535 questions.

    \item \textbf{GPQA}~\citep{gpqa} consists of graduate-level scientific QA questions designed to test expert-level scientific reasoning. We evaluate the medically related Genetics and Molecular Biology subsets, which contain 20 and 83 questions, respectively, for a total of 103 questions.
\end{itemize}

\subsection{Training Setup and Hyperparameters}
\label{apdx:exp:training}
Our implementation is built on Verl~\citep{verl} and vLLM~\citep{vllm}. Since proposer and reasoner rollouts are executed alternately, we disable GPU colocation in Verl and use Ray~\citep{ray} to repartition computational resources. Our final setup uses six 8-GPU nodes, with GPUs allocated to the proposer and reasoner in a 2:4 ratio, which yields the best runtime performance in our experiments.

Before self-play, we construct a small warm-start dataset as described in Section~\ref{apdx:data}. For fair comparison, all self-play methods use this same warm-start dataset for initialization. We warm-start the proposer and reasoner separately for 5 epochs with learning rate $1\times10^{-5}$. For self-play RL, we optimize the proposer and reasoner alternately with GRPO for $T=15$ iterations. Each iteration contains $U_P=10$ proposer-update steps and $U_R=10$ reasoner-update steps. During proposer rollouts, each document produces $N_p=4$ proposer candidates, and each candidate is answered by the reasoner $N_{\mathrm{acc}}=4$ times to estimate learnability. During reasoner updates, we sample $N_r=8$ responses per selected question. Full hyperparameters are listed in Table~\ref{tab:training_hyperparameters}.

\begin{table}[!htbp]
\centering
\caption{Key hyperparameters.}
\label{tab:training_hyperparameters}
\small
\setlength{\tabcolsep}{3.8pt}
\renewcommand{\arraystretch}{1.08}
\begin{tabular}{@{}lclclc@{}}
\toprule
\multicolumn{6}{c}{\textbf{SFT Warm-Start}} \\
\cmidrule(lr){1-6}
Epochs & 5 & Batch size & 128 & Max response len. & 8,192 \\
Learning rate & $1{\times}10^{-5}$ & LR schedule & Cosine & LR warmup ratio & 0.1 \\
\addlinespace[2pt]
\multicolumn{6}{c}{\textbf{Self-Play RL}} \\
\cmidrule(lr){1-6}
\multicolumn{2}{c}{Optimization} & \multicolumn{2}{c}{Rollout} & \multicolumn{2}{c}{Others} \\
\cmidrule(lr){1-2}\cmidrule(lr){3-4}\cmidrule(lr){5-6}
Learning rate & $1{\times}10^{-6}$ & Proposer samples $N_p$ & 4 & Invalid penalty $r^{\mathrm{format}}$ & -0.1 \\
Train batch & 128 & Answer samples $N_{\mathrm{acc}}$ & 4 & Learnability weight & 1.0 \\
Proposer steps $U_P$ & 10 & Reasoner samples $N_r$ & 8 & Rubric weight $\lambda$ & 0.2 \\
Reasoner steps $U_R$ & 10 & Max input len. & 4,096 & Number of knowledge points $M$ & 4 \\
Clip upper & 0.28 & Max response len. & 8,192 &  &  \\
\bottomrule
\end{tabular}
\end{table}

\section{Additional Experimental Results}

\begin{table*}[!htbp]
\centering
\caption{Full results for the main self-play comparisons.}
\label{tab:seed_robustness}
\normalsize
\setlength{\tabcolsep}{2.0pt}
\renewcommand{\arraystretch}{1.18}
\resizebox{0.95\textwidth}{!}{%
\begin{tabular}{@{}lccccccc>{\columncolor{AvgHighlight}}c@{}}
\toprule
\multirow{2}{*}{\textbf{Method}}
& \multicolumn{4}{c}{\textbf{Expert Medical Reasoning}}
& \multicolumn{1}{c}{\textbf{Clin. Case}}
& \multicolumn{2}{c}{\textbf{Science QA}}
& \cellcolor{white} \\
\cmidrule(lr){2-5}
\cmidrule(lr){6-6}
\cmidrule(lr){7-8}
& MedXpert & MB-OP4 & MB-OP5 & MedQA
& LiveClin & MMLU. & GPQA & Avg. \\
\midrule

R-Zero
& 19.7\,$\pm$\,2.3
& 72.9\,$\pm$\,0.5
& 68.6\,$\pm$\,2.5
& 83.5\,$\pm$\,0.4
& 66.0\,$\pm$\,1.5
& 73.3\,$\pm$\,0.4
& 52.9\,$\pm$\,2.6
& 62.4 \\
\quad 95\% CI
& [16.8, 21.6]
& [71.8, 74.2]
& [65.1, 71.4]
& [82.7, 84.1]
& [64.8, 69.1]
& [72.6, 74.1]
& [50.0, 56.5]
& --- \\

\addlinespace[3pt]
SPICE
& 21.8\,$\pm$\,2.2
& 74.2\,$\pm$\,1.0
& 70.0\,$\pm$\,2.0
& 83.2\,$\pm$\,0.4
& 67.1\,$\pm$\,1.2
& 75.6\,$\pm$\,0.2
& 54.2\,$\pm$\,1.6
& 63.7 \\
\quad 95\% CI
& [17.1, 24.3]
& [72.8, 75.2]
& [66.3, 73.1]
& [82.4, 83.9]
& [65.1, 69.9]
& [74.9, 75.9]
& [51.8, 57.1]
& --- \\

\addlinespace[3pt]
DARC
& 23.5\,$\pm$\,1.0
& 75.7\,$\pm$\,1.0
& 69.8\,$\pm$\,1.4
& 84.8\,$\pm$\,0.5
& 70.5\,$\pm$\,0.7
& 75.3\,$\pm$\,0.1
& 55.2\,$\pm$\,1.3
& 65.0 \\
\quad 95\% CI
& [22.0, 25.4]
& [73.6, 77.5]
& [66.2, 73.5]
& [83.1, 86.7]
& [68.7, 72.8]
& [74.3, 76.4]
& [53.6, 58.4]
& --- \\

\midrule
\textbf{\oursolution{}}
& \textbf{25.1\,{\boldmath$\pm$}\,1.2}
& \textbf{76.9\,{\boldmath$\pm$}\,1.4}
& \textbf{70.1\,{\boldmath$\pm$}\,1.8}
& \textbf{85.4\,{\boldmath$\pm$}\,0.5}
& \textbf{72.3\,{\boldmath$\pm$}\,0.8}
& \textbf{75.9\,{\boldmath$\pm$}\,0.3}
& \textbf{57.6\,{\boldmath$\pm$}\,1.0}
& \textbf{66.2} \\
\quad 95\% CI
& [23.5, 27.9]
& [74.8, 79.9]
& [67.4, 73.4]
& [84.6, 86.0]
& [70.9, 74.0]
& [75.0, 76.8]
& [55.8, 59.8]
& --- \\

\bottomrule
\end{tabular}%
}
\end{table*}

\paragraph{Robustness of Performance Gains and Confidence Intervals.}
\label{apdx:seed_robustness}
Table~\ref{tab:seed_robustness} reports mean accuracies and standard
deviations over three training seeds. Each checkpoint is evaluated
with eight responses per question (24 across the three checkpoints);
95\% CIs are estimated by percentile bootstrapping over training
runs and evaluation questions. \oursolution{} achieves the highest mean accuracy on all seven benchmarks, with an average accuracy of 66.2\%, compared with 65.0\% for DARC, 63.7\% for SPICE, and 62.4\% for R-Zero. The reported confidence intervals for \oursolution{} lie entirely above those for SPICE on MedQA and LiveClin, and above those for R-Zero on MedXpert, MB-OP4, MedQA, LiveClin, and MMLU-Pro. Although the intervals overlap in the remaining comparisons, the consistent advantage in mean accuracy across all seven benchmarks provides empirical support for the effectiveness of \oursolution{} in improving medical reasoning under the three-run evaluation.

\begin{table*}[!htbp]
\centering
\caption{Comparison with GRPO on a fixed synthetic question pool.}
\label{tab:static_synthetic}
\normalsize
\setlength{\tabcolsep}{2.0pt}
\renewcommand{\arraystretch}{1.18}
\resizebox{0.95\textwidth}{!}{%
\begin{tabular}{@{}lccccccc>{\columncolor{AvgHighlight}}c@{}}
\toprule
\multirow{2}{*}{\textbf{Method}}
& \multicolumn{4}{c}{\textbf{Expert Medical Reasoning}}
& \multicolumn{1}{c}{\textbf{Clin. Case}}
& \multicolumn{2}{c}{\textbf{Science QA}}
& \cellcolor{white} \\
\cmidrule(lr){2-5}
\cmidrule(lr){6-6}
\cmidrule(lr){7-8}
& MedXpert & MB-OP4 & MB-OP5 & MedQA
& LiveClin & MMLU. & GPQA & Avg. \\
\midrule
Static Synthetic Data + GRPO
& 22.6 & 75.0 & 69.5 & 84.1
& 66.9 & 75.6 & 55.3 & 64.1 \\
\textbf{\oursolution{}}
& \textbf{25.1} & \textbf{76.9} & \textbf{70.1} & \textbf{85.4}
& \textbf{72.3} & \textbf{75.9} & \textbf{57.6} & \textbf{66.2} \\
\bottomrule
\end{tabular}%
}
\end{table*}

\paragraph{Effectiveness of the Self-Play Training Paradigm.}
\label{apdx:static_synthetic}
To assess adaptive self-play relative to a fixed-data training pipeline, we use DeepSeek-V3.2~\citep{dpsk-v32} to generate approximately 19K question--answer pairs from the same 4,281 case reports. This synthetic pool remains fixed throughout GRPO training. We match the warm-start reasoner, the number of training pairs, and the reasoner GRPO update budget to \oursolution{}. As shown in Table~\ref{tab:static_synthetic}, \oursolution{} outperforms the static baseline on all seven benchmarks, achieving an average accuracy of 66.2\% versus 64.1\%. This comparison supports the effectiveness of adaptive self-play under the matched reasoner update budget. A plausible explanation is that adapting question generation to the evolving reasoner helps maintain informative response groups: questions that elicit uniformly correct or incorrect responses provide no correctness-driven group-relative learning signal under binary rewards.

\begin{table*}[!htbp]
\centering
\caption{Continuing medical RL post-training with \oursolution{}.}
\label{tab:medical_rl_continuation}
\normalsize
\setlength{\tabcolsep}{2.0pt}
\renewcommand{\arraystretch}{1.18}
\resizebox{0.95\textwidth}{!}{%
\begin{tabular}{@{}lccccccc>{\columncolor{AvgHighlight}}c@{}}
\toprule
\multirow{2}{*}{\textbf{Method}}
& \multicolumn{4}{c}{\textbf{Expert Medical Reasoning}}
& \multicolumn{1}{c}{\textbf{Clin. Case}}
& \multicolumn{2}{c}{\textbf{Science QA}}
& \cellcolor{white} \\
\cmidrule(lr){2-5}
\cmidrule(lr){6-6}
\cmidrule(lr){7-8}
& MedXpert & MB-OP4 & MB-OP5 & MedQA
& LiveClin & MMLU. & GPQA & Avg. \\
\midrule
Fleming-R1-7B
& 19.1 & 57.8 & 54.1 & 75.1
& 60.2 & 69.3 & 48.7 & 54.9 \\
\quad $+$ \oursolution{}
& \textbf{20.2} & \textbf{61.6} & \textbf{57.9} & \textbf{76.5}
& \textbf{65.3} & \textbf{69.6} & \textbf{50.2} & \textbf{57.3} \\
\bottomrule
\end{tabular}%
}
\end{table*}

\paragraph{Compatibility with Medical RL Post-Training.}
\label{apdx:medical_rl_continuation}
We next investigate whether \oursolution{} can further improve a reasoner that has already undergone medical RL post-training. We initialize the reasoner from Fleming-R1-7B~\citep{fleming-r1} and continue training with \oursolution{} on the clinical case-report corpus, using the same Avg@8 evaluation protocol. As shown in Table~\ref{tab:medical_rl_continuation}, average accuracy increases from 54.9\% to 57.3\%, a gain of 2.4 percentage points. Performance improves on all seven benchmarks, with the largest gain on LiveClin, from 60.2\% to 65.3\% (+5.1 percentage points). This continuation experiment shows that \oursolution{} provides additional gains for the evaluated medical RL model, supporting its use as a subsequent training stage after medical RL post-training.

\begin{figure*}[!htpb]
  \centering
  \begin{subfigure}[t]{0.32\textwidth}
    \centering
    \includegraphics[width=\linewidth]{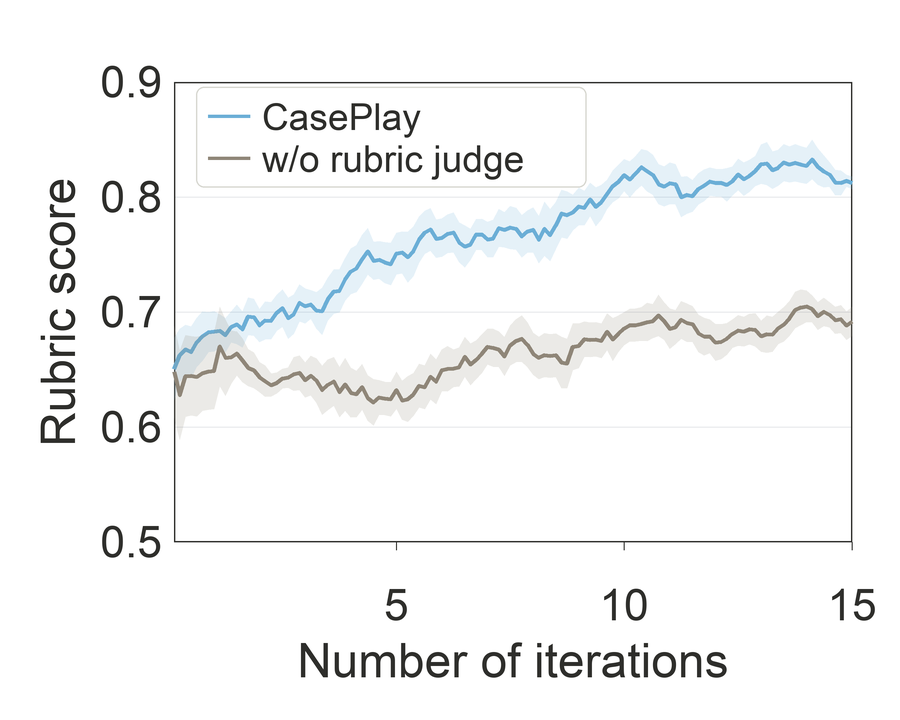}
      \vspace{-3mm}
    \caption{Clue integration}
    \label{fig:rubric_curve_clue}
  \end{subfigure}
  \hfill
  \begin{subfigure}[t]{0.32\textwidth}
    \centering
    \includegraphics[width=\linewidth]{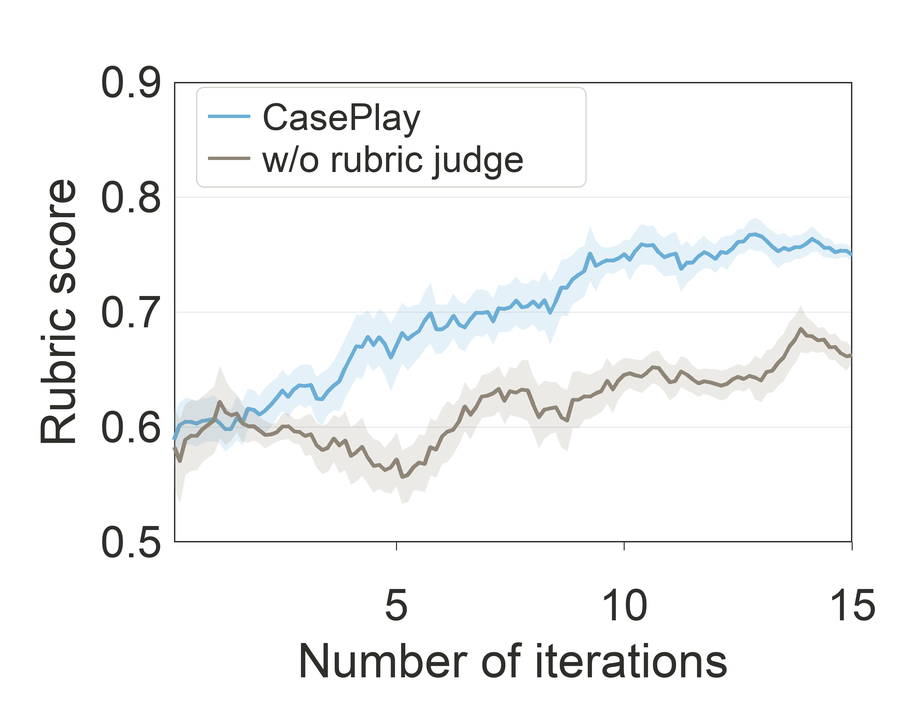}
      \vspace{-3mm}
    \caption{Reasoning depth}
    \label{fig:rubric_curve_multihop}
  \end{subfigure}
  \hfill
  \begin{subfigure}[t]{0.32\textwidth}
    \centering
    \includegraphics[width=\linewidth]{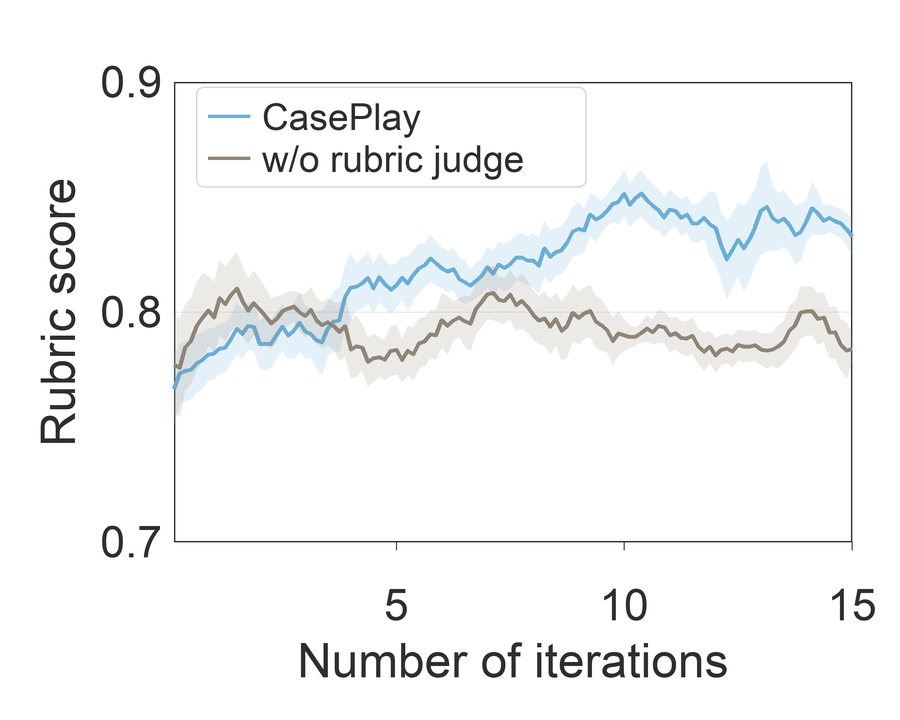}
      \vspace{-3mm}
    \caption{Option quality}
    \label{fig:rubric_curve_option}
  \end{subfigure}
  \caption{Training curves of rubric scores.}
    \vspace{-3mm}
  \label{fig:rubric_dimensions}
\end{figure*}

\paragraph{Quality Analysis of Generated Questions.}
\label{apdx:ablation_analysis}
We next examine how the rubric scores of generated questions evolve during training. Figure~\ref{fig:rubric_dimensions} compares the three rubric scores throughout training with and without rubric-based feedback. Incorporating this feedback improves all three dimensions, with the most pronounced gains in clue integration and reasoning depth. This pattern suggests that the rubric encourages the proposer to construct questions requiring the integration of multiple clinical clues and multistep reasoning, extending the training signal beyond estimated difficulty. Improvements in option quality are more modest. One plausible explanation is that constructing discriminative distractors depends more heavily on the proposer's parametric medical knowledge, while clue integration and reasoning depth can be influenced more directly through the selection and composition of case evidence. The variant without rubric feedback also improves at some stages, but its scores remain lower and less stable. These results indicate that learnability-based feedback alone provides weaker control over the quality of generated questions.

\paragraph{Impact of Knowledge-point Selection Strategy.}
\label{apdx:exp:kg_selection}

We compare three strategies for selecting knowledge points during question generation: \emph{random}, \emph{round-robin}, and \emph{accuracy-priority}. For a document $d$ with extracted knowledge points $K(d)=\{k_1,\ldots,k_M\}$, the \emph{random} strategy uniformly samples one knowledge point from $K(d)$ at each rollout. The \emph{round-robin} strategy selects knowledge points cyclically, as described in Eq.~(\ref{eq:round_robin_selector}). The \emph{accuracy-priority} strategy samples knowledge points according to the current reasoner's average answer accuracy on questions generated from each knowledge point. Specifically, for knowledge point $k_m$, its sampling probability is defined by applying a softmax over the learnability scores,
$p(k_m \mid d)=\mathrm{softmax}_{k_m\in K(d)}(1-2|\bar{a}_m-0.5|)$, where $\bar{a}_m$ is the current reasoner's average accuracy on questions associated with $k_m$.

\begin{wraptable}[9]{r}{0.48\columnwidth}
\vspace{-1.2em}
\centering
\small
\caption{Impact of knowledge-point selection strategies.}
\label{tab:kg_selection}
\setlength{\tabcolsep}{4pt}
\begin{tabular}{@{}lcc@{}}
\toprule
Strategy & DCR $\uparrow$ & Avg. Acc. $\uparrow$ \\
\midrule
Random & 0.40 & 65.7 \\
Acc.-priority & 0.32 & 64.1 \\
Round-robin & \textbf{0.42} & \textbf{66.2} \\
\bottomrule
\end{tabular}
\vspace{-1.2em}
\end{wraptable}

As shown in Table~\ref{tab:kg_selection}, \emph{round-robin} achieves the best performance, with the highest document coverage ratio (0.42) and the highest average accuracy (66.2). The \emph{random} strategy is competitive, but its stochastic selection can repeatedly sample the same knowledge points, leading to slightly lower coverage and downstream performance. In contrast, \emph{accuracy-priority} substantially reduces document coverage (0.32) and achieves the lowest average accuracy. Since each document is sampled only a limited number of times during self-play, prioritizing knowledge points based on noisy intermediate accuracy estimates can over-concentrate generation on a small subset of knowledge points, reducing within-document diversity and weakening the effectiveness of self-play RL. We therefore adopt \emph{round-robin} as the default strategy in \oursolution{}.

\section{Limitations}
\label{apdx:limitations}
Despite the strong empirical gains of \oursolution{}, our current implementation extracts a fixed set of knowledge points before self-play training and keeps them unchanged throughout RL. This design provides a stable and controllable basis for knowledge-conditioned generation, which helps reduce redundant question generation and improve training efficiency. However, it also limits the framework's ability to adapt its knowledge-point coverage as the proposer and reasoner improve. Future work could explore online discovery, pruning, and refinement of knowledge points when learning from unlabeled medical corpora. Additionally, our experiments focus on text-grounded case reports and text-only benchmark settings, while many medical corpora contain images that provide important clinical evidence. Extending \oursolution{} to multimodal self-play is therefore an important direction for improving its practical utility in medical reasoning. Such an extension could allow the framework to generate and evaluate questions grounded in both textual and visual evidence, making self-play learning better aligned with real clinical case reasoning.

\section{Broader Impacts}
\label{apdx:broder_impacts}
This work aims to improve medical reasoning models by learning from unlabeled clinical case reports. Potential benefits include reducing reliance on costly expert annotations and supporting research on medical education and clinical question answering. However, the work also has risks. Models trained with this framework may produce incorrect or overconfident medical answers, and case reports may contain biases because they often describe unusual or publication-worthy cases rather than representative patient populations. There are also privacy concerns when using medical text, even if the data are publicly available, and stronger medical reasoning models could be misused to generate misleading health information. We view \oursolution{} as a research framework, not a deployable clinical decision-making system. Any downstream use should require de-identified and properly licensed data, careful factuality and bias evaluation, clear communication of limitations, and human clinical oversight.

\end{document}